\documentclass{article}
\usepackage{iclr2027_conference,times}
\iclrfinalcopy                                     
\usepackage{bibunits}
\usepackage[hidelinks]{hyperref}
\usepackage{url}
\usepackage{booktabs}
\usepackage{graphicx}
\usepackage{xcolor}
\usepackage{colortbl}
\usepackage{placeins}
\usepackage{float}
\usepackage{flafter}
\usepackage{caption}
\usepackage{multirow}
\usepackage{wrapfig}
\usepackage{amsmath,amssymb,mathtools}
\usepackage{algorithm}
\usepackage{algpseudocode}

\newcommand{\R}{\mathbb{R}}
\newcommand{\Q}{\mathcal{Q}}
\newcommand{\V}{\mathcal{V}}
\newcommand{\cB}{\mathcal{B}}
\newcommand{\cD}{\mathcal{D}}

\newcommand{\norm}[1]{\left\lVert #1 \right\rVert}
\newcommand{\abs}[1]{\left\lvert #1 \right\rvert}
\newcommand{\tome}{ToMe}
\newcommand{\fastv}{FastV}
\newcommand{\visionzip}{VisionZip}
\newcommand{\framefusion}{FrameFusion}

\newcommand{\sttm}{STTM}
\newcommand{\flashvid}{FlashVID}
\newcommand{\learnpruner}{LearnPruner}
\newcommand{\venueyear}[2]{{\fontsize{6.3pt}{7pt}\selectfont\color{black!55}(#1-#2)}}
\newcommand{\venueyearlink}[3]{\hyperlink{cite.#3}{\venueyear{#1}{#2}}}
\definecolor{oursburgundy}{HTML}{DCEEFF}
\definecolor{lightblue}{HTML}{D6EAF8}

\newcommand{\tablebodyformat}{%
  \fontsize{7.59pt}{9.71pt}\selectfont
  \renewcommand{\arraystretch}{1.0}%
  \setlength{\extrarowheight}{0.5pt}%
  \setlength{\tabcolsep}{1.5pt}%
}

\title{FoveaTree: Information-Density Trees for\\
Query-Guided Visual Token Pruning in Video Large Language Models}

\author{
Zihan Meng\textsuperscript{1},
Weidong Chen\textsuperscript{1,\thanks{Corresponding author.}},
Dexiang Hong\textsuperscript{1},
Xinyan Liu\textsuperscript{2},
Zhendong Mao\textsuperscript{1}\\
\textsuperscript{1}University of Science and Technology of China\\
\textsuperscript{2}Harbin Institute of Technology
}

\begin{document}
\addtocontents{toc}{\protect\setcounter{tocdepth}{-1}}
\maketitle

\begin{abstract}
Video large language models (Video LLMs) have demonstrated strong multimodal understanding across a range of tasks, including visual question answering and video captioning. Yet, encoding each frame into hundreds of visual tokens makes inference over temporal visual content expensive. Existing methods reduce this cost by merging similar tokens or using text-to-vision attention to select task-relevant tokens. Nevertheless, these methods still face two main challenges: (1) \textbf{Hierarchy-blind Pruning}: many similarity-based methods rely on pairwise similarity or training a lightweight neural network to predict importance scores without explicitly considering the regional compressibility of a frame, which can misguide the merging or removal of critical evidence. (2) \textbf{Attention Drift}: position-dependent effects from Rotary Position Embedding (RoPE), which encodes token positions in the decoder, together with preferences for nearby or recent tokens and attention sinks can shift query-to-vision attention away from task relevance, causing attention-based pruning to underweight and discard task-relevant visual tokens. Motivated by these observations, we propose FoveaTree, a two-stage, training-free visual-token pruning framework for efficient video inference. The first stage estimates regional compressibility from the feature distortion incurred when merging a region to construct a quadtree-derived density map, which guides the dropping, merging, and retention of visual regions. Based on the surviving units and inherited tree topology, the second stage introduces an efficient tree-structured pruning mechanism in the middle layers of the LLM to mitigate Attention Drift and further reduce low-relevance visual content using text-to-vision attention. We validate FoveaTree under different Video LLM architectures, i.e., InternVL3.5-8B and Qwen3-VL-8B, on five video question-answering benchmarks and one captioning benchmark with token budgets of 50\%, 30\%, and 10\%; our method consistently achieves overall state-of-the-art performance across all evaluated settings, demonstrating a superior accuracy-efficiency trade-off.\footnote{Code will be released in the final version of the paper.}
\end{abstract}

\section{Introduction}

Recent video large language models (Video LLMs)~\citep{wang2025internvl35,bai2025qwen3vl} exhibit strong video understanding, delivering competitive performance on tasks such as video question answering and video captioning~\citep{li2025llavaonevision,bai2025qwen25vl}. A Video LLM contains a vision encoder, a modality projector, and an LLM~\citep{liu2023llava,lin2024videollava,li2024llamavid}; the encoder represents each frame as patch tokens, which the projector maps into the language embedding space~\citep{dosovitskiy2021vit,radford2021clip,zhai2023siglip,oquab2023dinov2,he2022mae,liu2021swin}. Since the visual sequence grows with video length and resolution~\citep{wang2024qwen2vl,xu2024llavauhd}, it inflates attention computation, prefill latency, and key-value cache memory, motivating visual-token pruning for practical inference~\citep{tu2024vlcache}.

\begin{figure*}[t]
    \centering
    \includegraphics[width=\textwidth]{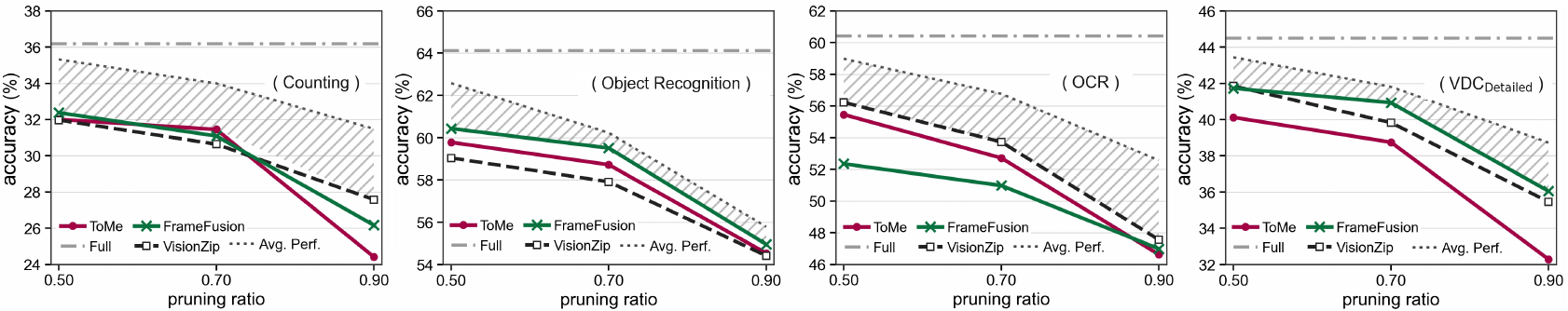}
    \caption{Illustration of Hierarchy-blind Pruning on fine-grained domains from VideoMME~\citep{fu2024videomme} and VDC~\citep{chai2025auroracap}. Avg. Perf. summarizes performance retention on the complete evaluation set and is projected onto each domain's full-model score, whereas the method curves report the measured domain-level performance. The gray regions show the resulting gap on OCR, counting, object recognition, and detailed captioning across pruning ratios.}
    \label{fig:preliminary-study}
\end{figure*}

Visual-token pruning lowers this cost~\citep{shang2024prumerge,endo2024feather,zou2025holov,xing2024pyramiddrop,arif2024hired,zhang2024fastervlm}. ToMe~\citep{bolya2023tome} and EViT~\citep{liang2022evit} merge or discard redundant tokens~\citep{ryoo2021tokenlearner,rao2021dynamicvit,fayyaz2022ats,marin2023tokenpooling}; VisionZip~\citep{yang2025visionzip}, FrameFusion~\citep{fu2025framefusion}, and STTM~\citep{hyun2025sttm} extend merging to cross-frame and spatiotemporal redundancy; and FastV~\citep{chen2024fastv}, MADTP~\citep{cao2024madtp}, and LearnPruner~\citep{takezoe2026learnpruner} select query-related tokens via text-to-vision attention~\citep{ye2025atpllava,zhang2024sparsevlm}. These methods shorten visual sequences but provide limited control over how evidence is organized~\citep{wen2025dart}.

\begin{wrapfigure}{r}{0.52\linewidth}
    \vspace{-8pt}
    \centering
    \includegraphics[width=\linewidth]{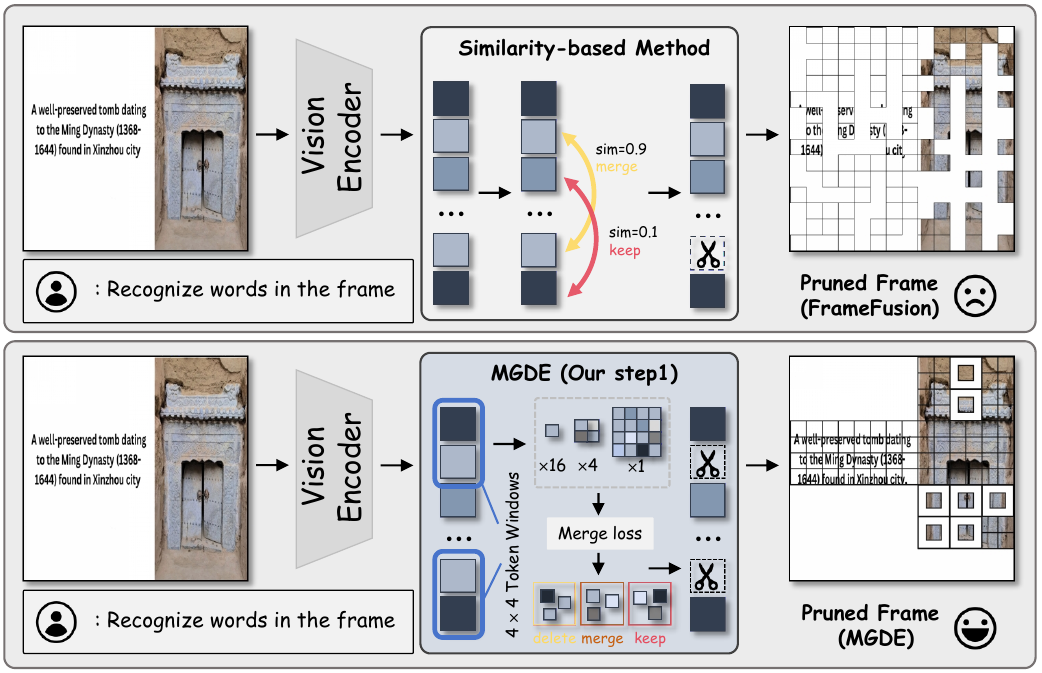}
    \caption{Architectural comparison of FrameFusion and MGDE. FrameFusion merges corresponding tokens across adjacent frames; MGDE instead constructs an explicit multiscale regional hierarchy.}
    \label{fig:pre-mdge}
    \vspace{-4pt}
\end{wrapfigure}

\begin{figure}[b]
    \centering
    \includegraphics[width=\linewidth]{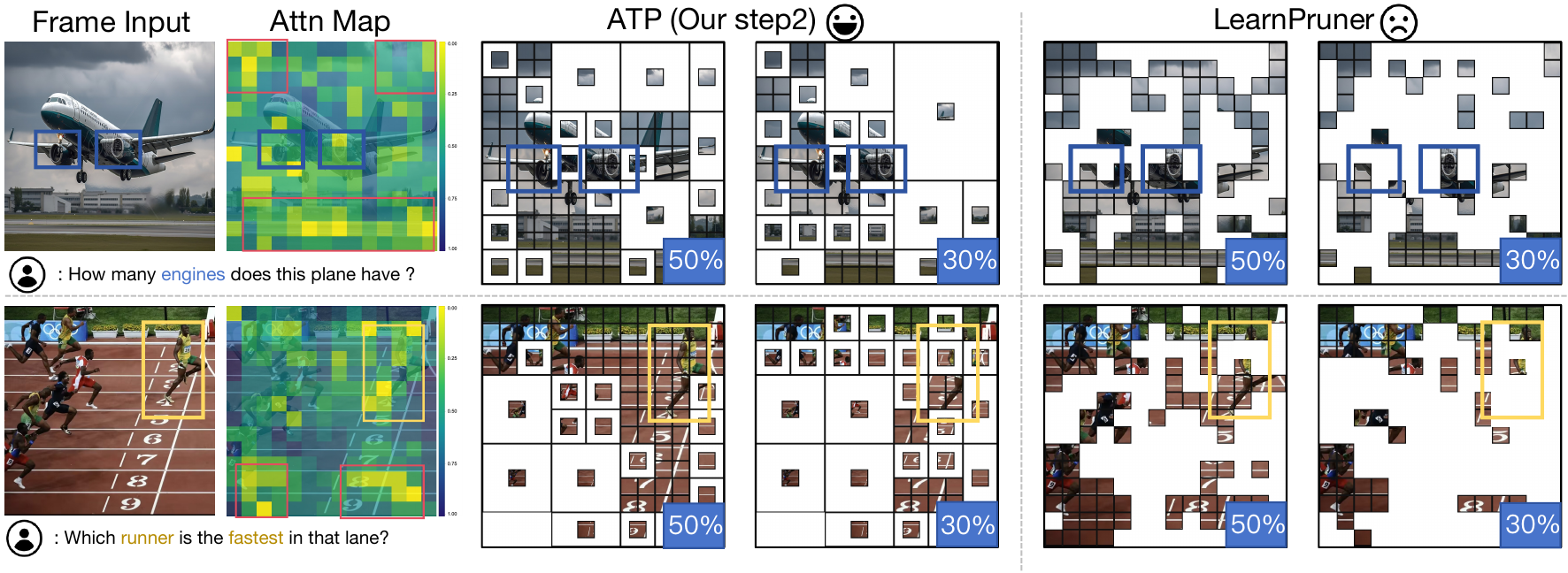}
    \caption{Visualization of our tree-based attention pruning and the conventional attention-based top-$k$ pruning used by LearnPruner~\citep{takezoe2026learnpruner} at 50\% and 30\% visual-token retention for two input frames. The comparison illustrates Attention Drift around peripheral and task-irrelevant regions (red boxes), while the tree structure improves robustness to this bias at both budgets.}
    \label{fig:pre-atp-viz}
\end{figure}

Fig.~\ref{fig:preliminary-study} compares overall and domain-level retention for ToMe, VisionZip, and FrameFusion. Avg. Perf. accounts for general performance degradation at each budget, whereas the method curves report measured domain performance. The gray regions therefore isolate additional losses of fine-grained evidence rather than a uniform decline of the model. The panels test small text (OCR), distinct instances (counting), local appearance and boundaries (object recognition), and multiple visual details within one description (VDC detailed captioning). All require coherent evidence across spatial scales, yet similarity methods make pairwise token-level decisions without representing how cells form nested regions. Their systematic domain deficit thus exposes a blind spot to regional hierarchy, which we term \textbf{Hierarchy-blind Pruning}. Fig.~\ref{fig:pre-mdge} illustrates this limitation with FrameFusion, a representative similarity-based method. FrameFusion computes cosine similarity only between spatially corresponding tokens in adjacent frames and performs cascaded merging, but it neither estimates within-frame regional compressibility nor constructs a nested spatial hierarchy.


Fig.~\ref{fig:pre-atp-viz} visualizes query-to-vision attention scores. \textbf{Attention Drift} appears as anomalously high responses at spatial boundaries or early visual-sequence positions despite weak query relevance, reflecting RoPE positional effects, nearby or recent-token preferences, and attention sinks~\citep{su2024roformer,xiao2024streamingllm,zhao2025attentiondebiasing,zhang2025d2pruner,liu2024lostmiddle}. Flat top-$k$ methods such as LearnPruner follow these peaks; as retention falls from 50\% to 30\%, query-relevant regions become increasingly fragmented or lost, whereas ATP uses the inherited tree to preserve relevant regions together with their context, keeping the surrounding visual evidence intact during pruning.

Motivated by these observations, we propose FoveaTree, a two-stage, training-free visual-token pruning framework. Merge-Gain Density Estimation (MGDE) builds a quadtree for each frame, estimates compressibility from merge-induced feature distortion, labels terminal regions by density, and performs drop, merge, and keep operations. Attention Tree Pruning (ATP) then uses middle-layer text-to-vision attention to prune the surviving hierarchy to the target budget per frame.

Together, the two stages preserve core regions more precisely, address the challenges above, and maintain spatial context during query-guided pruning. We validate FoveaTree under different Video LLM architectures, InternVL3.5-8B and Qwen3-VL-8B, across five video question-answering benchmarks and one captioning benchmark at 50\%, 30\%, and 10\% visual-token retention. Across all evaluated settings, FoveaTree achieves overall state-of-the-art performance and demonstrates a superior accuracy-efficiency trade-off across both backbones and all retention budgets.

In summary, our main contributions are as follows:
\begin{itemize}
    \item We propose FoveaTree, a two-stage, training-free visual-token pruning framework for efficient Video LLM inference. By combining compressibility-guided spatial pruning with query-guided tree pruning within the LLM, FoveaTree efficiently preserves task-relevant visual evidence at low visual-token retention without additional training or labels.
    \item We introduce MGDE, which builds a multiscale hierarchy and density map from merge distortion and gain, then adapts pruning to regional compressibility by dropping, merging, or retaining tokens. We also introduce ATP to propagate attention scores through the inherited quadtree, improving robustness to Attention Drift and preserving task-relevant regions.
    \item We conduct experiments on five video question-answering benchmarks and one captioning benchmark to evaluate FoveaTree across different Video LLM architectures, while component ablations on InternVL3.5-8B assess each proposed module. Notably, at 30\% visual-token retention on InternVL3.5-8B, FoveaTree achieves 100.62\% performance of the full model, with a $1.24\times$ inference speedup and 53.2\% lower TFLOPs.
\end{itemize}

\section{Method}
\label{sec:method}

\subsection{Preliminary}

Let $M_\theta$ denote a Video LLM with a vision encoder, multimodal projector, $L$-layer language decoder, and output projection. Given a video $v=\{v_t\}_{t=1}^{T}$ and query $q$, the encoder and projector map each frame to an LLM-aligned grid $X_t=\{x_{t,p}\mid p\in\mathcal P\}$, where $\mathcal P=\{1,\ldots,H\}\times\{1,\ldots,W\}$ and $x_{t,p}\in\R^d$. The visual tokens are concatenated with the embedded query, and the decoder predicts $p_\theta(\cdot\mid v,q,y_{<r})=\operatorname{softmax}(f_\theta(\cdot\mid v,q,y_{<r}))$ before sampling the next token $y_r$. The original visual prefix contains $N=T|\mathcal P|=THW$ tokens, so prefill computation and KV-cache memory grow with $N$. Given a retention ratio $0<\eta\leq1$, we define the nominal per-frame budget as $B=\lceil\eta HW\rceil$ from the original visual grid. We denote projected features by $x_{t,p}$ and shallow vision features by $z_{t,p}$; MGDE uses the latter for spatial estimation before decoder prefill in every frame.

\begin{figure}[t!]
    \centering
    \includegraphics[width=\linewidth]{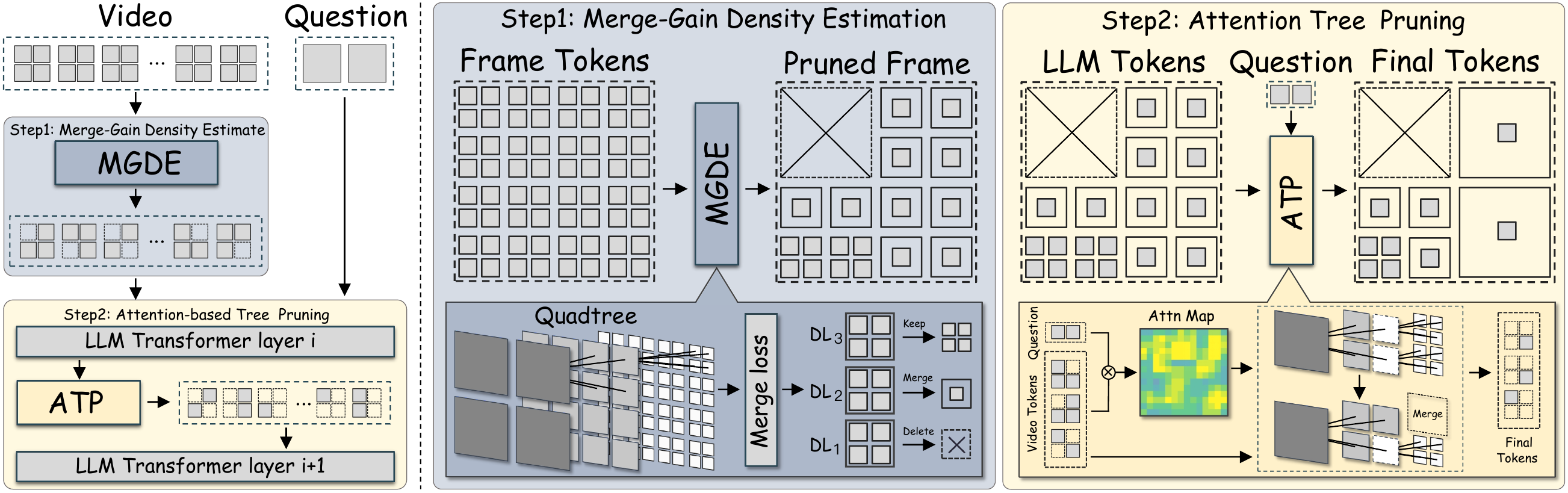}
    \caption{FoveaTree overview. MGDE builds a regional hierarchy and compresses visual tokens before decoder prefill; ATP then uses text-to-vision attention to prune the surviving hierarchy.}
    \label{fig:framework}
\end{figure}

\subsection{FoveaTree Framework}
\label{sec:framework}

As shown in Fig.~\ref{fig:framework}, FoveaTree reduces visual tokens in two stages: regional compression before decoder prefill and query-guided pruning within the decoder. MGDE first uses layer-0 vision features to construct a quadtree forest $\mathcal T_t$ per frame. The sizes of its terminal regions then define an operational density map $\mathcal M_t$, which guides dropping, merging, and keeping of projected visual tokens, producing a compact unit set $\V$ whose elements retain their spatial support and ancestry in $\mathcal T_t$. At an intermediate decoder layer, ATP assigns text-to-vision attention scores to $\V$ and propagates them over the surviving branches of $\mathcal T_t$. A top-down cut selects the resolution of each region under the per-frame budget during decoder processing, linking prefill compression with query-guided pruning.

\subsection{Merge-Gain Density Estimation}
\label{sec:mgde}

\paragraph{Motivation.}
As discussed above, similarity-based pruning can merge or discard tokens from informative regions because it does not explicitly represent the spatial hierarchy of a frame, and learned selectors require training and repeated model evaluation. MGDE provides a training-free alternative by measuring the feature distortion caused by regional merging. It uses this signal to construct a multiscale partition, then derives a density map that assigns finer or coarser token resolution according to the terminal region size before the visual tokens enter the language decoder for prefill.

\begin{wrapfigure}{r}{0.52\linewidth}
    \vspace{-8pt}
    \centering
    \includegraphics[width=\linewidth]{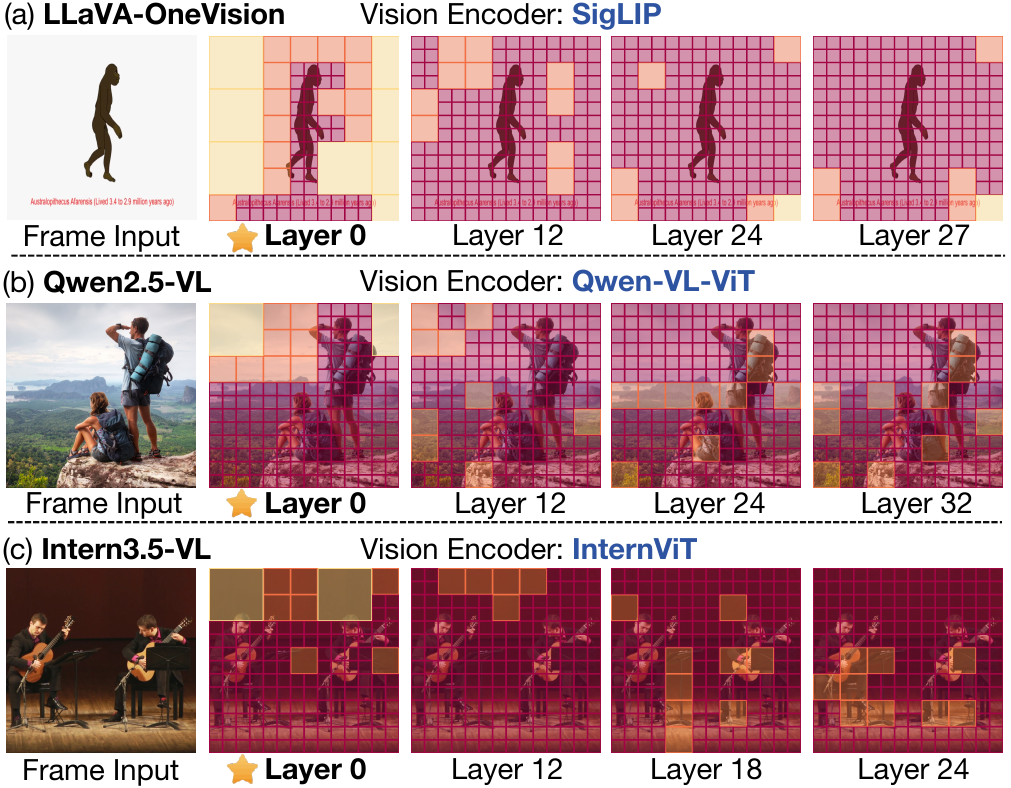}
    \caption{MGDE density maps from three vision encoders. Layer-0 features preserve texture and boundary cues, yielding regional hierarchies, whereas deeper features obscure spatial details.}
    \label{fig:mdge-layer}
    \vspace{-4pt}
\end{wrapfigure}

\paragraph{Regional compressibility.}
MGDE computes regional compressibility from shallow features $z_{t,p}$. Let $\bar z_{t,p}=z_{t,p}/(\norm{z_{t,p}}_2+\epsilon)$, where $\epsilon>0$ is a numerical constant. For a quadtree region $R$, let $\cB(R)=R\cap\mathcal P$ be its valid grid cells. We define the distortion of representing $R$ with one token as $\cD_t(R)=\abs{\cB(R)}-\left\lVert\sum_{p\in\cB(R)}\bar z_{t,p}\right\rVert_2$. If $R^{(1)},\ldots,R^{(4)}$ are its four quadrants, the gain from preserving their separate representations is $G_t(R)=\cD_t(R)-\sum_{k=1}^{4}\cD_t\bigl(R^{(k)}\bigr)$. Low distortion indicates that a region is suitable for coarse representation, whereas a large merge gain indicates greater feature variation within the region and local structure that should remain separated.
To determine which vision features best support MGDE, we extract features from multiple encoder layers and use each of them for regional analysis. Fig.~\ref{fig:mdge-layer} shows a consistent pattern across three vision encoders: layer-0 features most clearly resolve frame complexity and its information-density hierarchy across the high-, medium-, and low-density regions, because they preserve local textures and boundaries whereas deeper features become more semantic and less spatially localized. Based on this pattern, MGDE uses layer-0 features to construct the density map for every frame.
\begin{figure}[t]
    \centering
    \includegraphics[width=\linewidth]{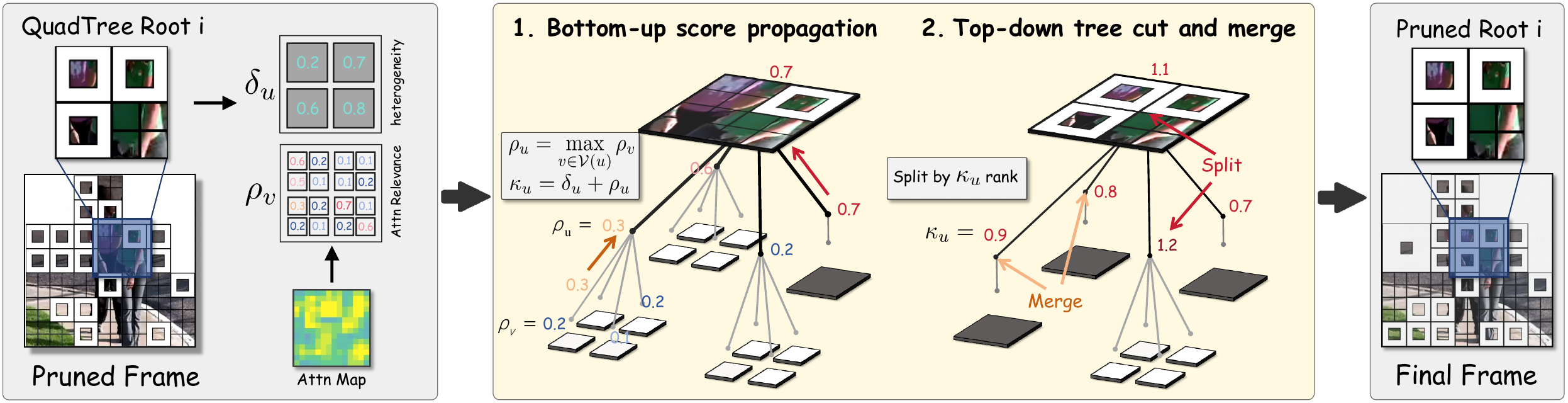}
    \caption{Visualization of ATP on one pruned video frame after MGDE. MGDE provides the surviving visual units and their inherited quadtree. ATP assigns text-to-vision attention scores to these units, propagates them to ancestor nodes, and performs top-down tree cutting and merging.}
    \label{fig:atp-tree}
\end{figure}

\paragraph{Density map and token construction.}
After padding, MGDE tiles each frame with non-overlapping $8\times8$ roots following the $8\rightarrow4\rightarrow2\rightarrow1$ hierarchy. It splits the highest-gain eligible region when $G_t(R)>\gamma$ while keeping at most $L_{\max}$ leaves across the forest $\mathcal T_t$, whose terminal regions form $\mathcal L_t$. MGDE then assigns each terminal region $R\in\mathcal L_t$ an operational density level according to its side length $s(R)$: regions with $s(R)\geq4$ receive $\mathrm{DL}_1$, $2\times2$ regions receive $\mathrm{DL}_2$, and single-cell regions receive $\mathrm{DL}_3$. These labels form the derived density map $\mathcal M_t$ and determine token construction: $\mathrm{DL}_1$ regions are dropped, projected features within each $\mathrm{DL}_2$ region are averaged into one token, and each $\mathrm{DL}_3$ cell is kept. Every resulting unit retains its source support, ancestry, and earliest raster position. The units, collected as $\V$ across frames, remain linked to $\mathcal T_t$; pooling stays within each frame and the original temporal order is preserved. ATP receives $\V$ and the restriction of $\mathcal T_t$ to branches with surviving units for subsequent query-guided pruning within the decoder.

\subsection{Attention Tree Pruning}
\label{sec:atp}

\paragraph{Motivation.}
\begin{wrapfigure}{r}{0.36\linewidth}
    \vspace{-8pt}
    \centering
    \includegraphics[width=\linewidth]{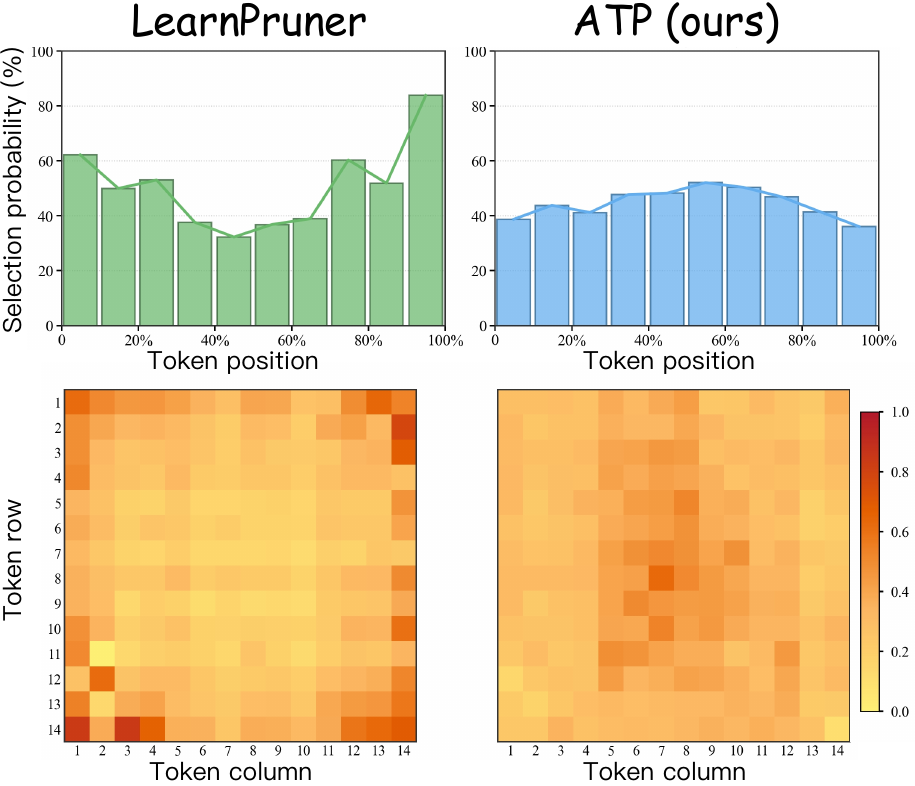}
    \caption{Token-selection frequencies of LearnPruner's top-$k$ pruning and ATP. ATP reduces peripheral selections and retains more tokens in the central region.}
    \label{fig:atp-heatmap}
    \vspace{-4pt}
\end{wrapfigure}
After MGDE, the surviving units retain a spatial hierarchy but can still contain content unrelated to the query. To examine how attention-based pruning distributes its selections, we run LearnPruner and ATP on Video-MME at the same visual-token retention ratio using the original query associated with each sample, then count how often the token at each spatial position is retained across frames. Fig.~\ref{fig:atp-heatmap} reports position-wise selection frequencies and heatmaps. Under Attention Drift, LearnPruner's flat attention top-$k$ tends to retain peripheral tokens, producing scattered peaks that can fragment relevant visual evidence. In contrast, ATP propagates the same attention scores through the inherited tree and selects spatially coherent branches, shifting retention toward central regions where the primary visual content of a frame is more frequently located. In this way, the tree structure lets ATP evaluate each branch within its multiscale spatial context, which is cheaper and more robust during pruning.

\begin{table*}[!t]
\centering
\begin{minipage}[t]{\textwidth}
\centering
\caption{InternVL3.5-8B results at 10\%, 30\%, and 50\% visual-token retention. Bold and underlined values mark the best and second-best compressed results; the gray line provides the baseline result.}
\label{tab:main-backbone-a}
{\scriptsize
\renewcommand{\arraystretch}{1.05}
\renewcommand{\multirowsetup}{\centering}
\setlength{\extrarowheight}{0.5pt}
\setlength{\tabcolsep}{0pt}
\resizebox{\linewidth}{!}{%
\begin{tabular}{>{\raggedright\arraybackslash}m{1.22in}|*{5}{>{\centering\arraybackslash}m{0.5435in}}|*{2}{>{\centering\arraybackslash}m{0.5435in}}|>{\centering\arraybackslash}m{0.465in}}
\hline
\multirow[c]{2}{1.2in}{\raggedright\textbf{Method}} & \multicolumn{5}{c|}{\textbf{Video QA}} & \multicolumn{2}{c|}{\textbf{Video Caption}} & \multirow[c]{2}{0.465in}{\centering\textbf{Avg.}} \\
\cline{2-8}
& \textbf{VideoMME} & \textbf{NExT-QA} & \textbf{MVBench} & \textbf{Valor32k} & \textbf{VNBench} & \textbf{VDC$_{\mathrm{Acc.}}$} & \textbf{VDC$_{\mathrm{Score}}$} & \\
\hline
\rowcolor{gray!15}
\mbox{InternVL3.5 (16,384 Tokens)} & 65.67 & 80.22 & 72.19 & 76.63 & 59.85 & 40.47 & 1.90 & 100.00\% \\
\hline
\multicolumn{9}{c}{\textit{Retain 8,192 Tokens ($\downarrow$ \textbf{50.0\%})}} \\
\hline
\tome{} \venueyearlink{ICLR}{2023}{bolya2023tome} & 62.64 & 76.48 & 69.02 & 73.14 & 57.18 & 38.67 & 1.82 & 95.52\% \\
\fastv{} \venueyearlink{ECCV}{2024}{chen2024fastv} & 62.05 & 75.79 & 68.27 & 72.51 & 56.96 & 38.45 & \underline{1.89} & 95.40\% \\
\visionzip{} \venueyearlink{CVPR}{2025}{yang2025visionzip} & 63.26 & 77.24 & 69.42 & 74.10 & 57.89 & 39.20 & 1.81 & 96.33\% \\
\framefusion{} \venueyearlink{ICCV}{2025}{fu2025framefusion} & 64.10 & 78.61 & 70.36 & 75.06 & 58.75 & 39.72 & 1.67 & 96.46\% \\
\sttm{} \venueyearlink{ICCV}{2025}{hyun2025sttm} & 65.23 & 79.75 & 72.03 & \textbf{76.44} & 59.76 & 40.46 & 1.85 & 99.35\% \\
\flashvid{} \venueyearlink{ICLR}{2026}{2026flashvid} & \textbf{65.70} & \underline{80.33} & \underline{72.09} & \underline{76.34} & \underline{59.95} & \underline{40.67} & \underline{1.89} & \underline{99.97\%} \\
\learnpruner{} \venueyearlink{ICLR}{2026}{takezoe2026learnpruner} & 64.87 & 78.91 & 71.30 & 75.48 & 59.19 & 39.70 & 1.83 & 98.25\% \\
\cellcolor{oursburgundy}\textbf{FoveaTree (Ours)} & \cellcolor{oursburgundy}\underline{65.41} & \cellcolor{oursburgundy}\textbf{80.53} & \cellcolor{oursburgundy}\textbf{72.25} & \cellcolor{oursburgundy}76.26 & \cellcolor{oursburgundy}\textbf{60.00} & \cellcolor{oursburgundy}\textbf{41.41} & \cellcolor{oursburgundy}\textbf{1.95} & \cellcolor{oursburgundy}\textbf{100.69\%} \\
\hline
\multicolumn{9}{c}{\textit{Retain 4,928 Tokens ($\downarrow$ \textbf{70.0\%})}} \\
\hline
\tome{} \venueyearlink{ICLR}{2023}{bolya2023tome} & 61.59 & 75.36 & 67.60 & 71.68 & 56.26 & 37.70 & 1.79 & 93.75\% \\
\fastv{} \venueyearlink{ECCV}{2024}{chen2024fastv} & 61.22 & 74.57 & 66.93 & 70.97 & 55.45 & 37.68 & 1.80 & 93.14\% \\
\visionzip{} \venueyearlink{CVPR}{2025}{yang2025visionzip} & 62.56 & 76.80 & 69.20 & 73.42 & 57.33 & 38.46 & 1.81 & 95.54\% \\
\framefusion{} \venueyearlink{ICCV}{2025}{fu2025framefusion} & 62.24 & 76.23 & 68.37 & 72.57 & 57.01 & 38.27 & 1.77 & 94.60\% \\
\sttm{} \venueyearlink{ICCV}{2025}{hyun2025sttm} & 64.33 & 79.07 & 70.96 & 75.44 & 58.90 & 39.57 & \underline{1.89} & 98.42\% \\
\flashvid{} \venueyearlink{ICLR}{2026}{2026flashvid} & \underline{65.12} & \textbf{80.06} & \underline{71.78} & \underline{76.28} & \textbf{59.64} & \underline{40.23} & \underline{1.89} & \underline{99.50\%} \\
\learnpruner{} \venueyearlink{ICLR}{2026}{takezoe2026learnpruner} & 63.66 & 77.76 & 69.91 & 74.52 & 58.13 & 39.11 & 1.83 & 96.86\% \\
\cellcolor{oursburgundy}\textbf{FoveaTree (Ours)} & \cellcolor{oursburgundy}\textbf{66.01} & \cellcolor{oursburgundy}\underline{80.04} & \cellcolor{oursburgundy}\textbf{72.49} & \cellcolor{oursburgundy}\textbf{76.34} & \cellcolor{oursburgundy}\underline{59.33} & \cellcolor{oursburgundy}\textbf{41.39} & \cellcolor{oursburgundy}\textbf{1.95} & \cellcolor{oursburgundy}\textbf{100.62\%} \\
\hline
\multicolumn{9}{c}{\textit{Retain 1,664 Tokens ($\downarrow$ \textbf{90.0\%})}} \\
\hline
\tome{} \venueyearlink{ICLR}{2023}{bolya2023tome} & 56.92 & 70.01 & 62.77 & 66.60 & 51.88 & 35.22 & 1.67 & 87.06\% \\
\fastv{} \venueyearlink{ECCV}{2024}{chen2024fastv} & 58.22 & 71.13 & 63.98 & 67.94 & 53.02 & 35.84 & 1.67 & 88.52\% \\
\visionzip{} \venueyearlink{CVPR}{2025}{yang2025visionzip} & 60.10 & 72.99 & 65.91 & 69.76 & 54.56 & 36.75 & 1.75 & 91.27\% \\
\framefusion{} \venueyearlink{ICCV}{2025}{fu2025framefusion} & 60.42 & 73.71 & 66.36 & 70.36 & 55.04 & 37.33 & 1.79 & 92.29\% \\
\sttm{} \venueyearlink{ICCV}{2025}{hyun2025sttm} & 63.58 & 77.92 & 70.20 & 74.21 & 57.82 & 39.33 & \underline{1.88} & 97.25\% \\
\flashvid{} \venueyearlink{ICLR}{2026}{2026flashvid} & \underline{63.96} & \underline{78.27} & \underline{70.76} & \underline{74.75} & \underline{58.35} & \underline{39.36} & 1.83 & \underline{97.37\%} \\
\learnpruner{} \venueyearlink{ICLR}{2026}{takezoe2026learnpruner} & 62.87 & 76.53 & 68.78 & 73.13 & 57.36 & 38.88 & 1.77 & 95.27\% \\
\cellcolor{oursburgundy}\textbf{FoveaTree (Ours)} & \cellcolor{oursburgundy}\textbf{65.41} & \cellcolor{oursburgundy}\textbf{79.98} & \cellcolor{oursburgundy}\textbf{72.04} & \cellcolor{oursburgundy}\textbf{75.96} & \cellcolor{oursburgundy}\textbf{59.25} & \cellcolor{oursburgundy}\textbf{40.02} & \cellcolor{oursburgundy}\textbf{1.89} & \cellcolor{oursburgundy}\textbf{99.37\%} \\
\hline
\end{tabular}%
}}
\end{minipage}\end{table*}

\paragraph{Bottom-up score propagation.}
Let $\Q$ be the question-token positions, $H_\ell$ the number of heads at decoder layer $\ell$, and $A^\ell_{qhv}$ the causal attention probability from question token $q\in\Q$ to visual unit $v\in\V$ in head $h$. ATP averages these probabilities over question tokens and heads to obtain $a_v$, then rescales $a_v$ within each frame between its empirical second and 98th percentiles and clips to $[0,1]$ to obtain the leaf relevance $\rho_v$. Each surviving unit is associated with the corresponding terminal node of the inherited MGDE quadtree, which connects its attention score to a spatial region. For an internal node $u$, let $\V(u)\subseteq\V$ denote its surviving descendant units. ATP processes the tree from the leaves to the root and assigns each internal node the strongest relevance among its descendants as the explicit regional relevance term that is subsequently consumed by the top-down tree-cut stage:
\begin{equation}
    a_v=\frac{1}{\abs{\Q}H_\ell}\sum_{q\in\Q}\sum_{h=1}^{H_\ell}A^\ell_{qhv},
    \qquad
    \rho_u=\max_{v\in\V(u)}\rho_v.
\end{equation}
This propagation converts token-level attention into region-level relevance without introducing learned parameters. The max operation assigns a branch the score of its strongest descendant, so a locally high score is evaluated together with the descendant's ancestral region. The top-down stage uses these propagated scores to determine the resolution retained for each branch under budget.

\paragraph{Top-down tree cut and merge.}
ATP combines propagated relevance with the heterogeneity of hidden states $h_v^\ell\in\R^d$ within each node. With $n_u=\abs{\V(u)}$, it computes the following scores:
\begin{equation}
    \delta_u=1-\frac{1}{n_u}\left\lVert\sum_{v\in\V(u)}\frac{h_v^\ell}{\norm{h_v^\ell}_2+\epsilon}\right\rVert_2,
    \qquad
    \kappa_u=\delta_u+\rho_u.
\end{equation}
The score $\kappa_u$ determines the split priority used to construct a budget-constrained frontier for each frame. If MGDE produces $M_t$ units for frame $t$, ATP targets $K_t=\min(B,M_t)$ outputs. It expands the tree in priority order and merges the residual siblings when a complete expansion would exceed the remaining budget, thereby returning exactly $K_t$ units. For a merged cluster $C\subseteq\V$, ATP attention-weights its hidden states and anchors the result at the highest-attention constituent's original RoPE position. After frame and raster reordering, preceding-layer caches remain unchanged, while subsequent layers populate fresh caches for the compressed prefix. The supplementary material further details the budget control and its boundary handling for exact per-frame output counts.

\FloatBarrier
\section{Experiments}
\label{sec:experiments}

\setcounter{topnumber}{1}
\begin{table*}[!t]
\centering
\begin{minipage}[t]{\textwidth}
\centering
\caption{Qwen3-VL-8B results at 10\%, 30\%, and 50\% visual-token retention. Bold and underlined values mark the best and second-best compressed results; the gray line provides the baseline result.}
\label{tab:main-backbone-b}
{\scriptsize
\renewcommand{\arraystretch}{1.05}
\renewcommand{\multirowsetup}{\centering}
\setlength{\extrarowheight}{0.5pt}
\setlength{\tabcolsep}{0pt}
\resizebox{\linewidth}{!}{%
\begin{tabular}{>{\raggedright\arraybackslash}m{1.22in}|*{5}{>{\centering\arraybackslash}m{0.5435in}}|*{2}{>{\centering\arraybackslash}m{0.5435in}}|>{\centering\arraybackslash}m{0.465in}}
\hline
\multirow[c]{2}{1.22in}{\raggedright\textbf{Method}} & \multicolumn{5}{c|}{\textbf{Video QA}} & \multicolumn{2}{c|}{\textbf{Video Caption}} & \multirow[c]{2}{0.465in}{\centering\textbf{Avg.}} \\
\cline{2-8}
& \textbf{VideoMME} & \textbf{NExT-QA} & \textbf{MVBench} & \textbf{Valor32k} & \textbf{VNBench} & \textbf{VDC$_{\mathrm{Acc.}}$} & \textbf{VDC$_{\mathrm{Score}}$} & \\
\hline
\rowcolor{gray!15}
\mbox{Qwen3-VL (5,753 Tokens)} & 66.15 & 83.43 & 67.49 & 78.27 & 58.63 & 45.28 & 2.31 & 100.00\% \\
\hline
\multicolumn{9}{c}{\textit{Retain 2,876 Tokens ($\downarrow$ \textbf{50.0\%})}} \\
\hline
\tome{} \venueyearlink{ICLR}{2023}{bolya2023tome} & 62.88 & 79.27 & 64.32 & 74.46 & 55.83 & 43.11 & 2.20 & 95.17\% \\
\fastv{} \venueyearlink{ECCV}{2024}{chen2024fastv} & 63.59 & 80.20 & 64.92 & 75.35 & 56.77 & 43.76 & 2.24 & 96.45\% \\
\visionzip{} \venueyearlink{CVPR}{2025}{yang2025visionzip} & 64.25 & 81.00 & 65.42 & 76.30 & 57.18 & 44.20 & 2.24 & 97.25\% \\
\framefusion{} \venueyearlink{ICCV}{2025}{fu2025framefusion} & 64.26 & 81.36 & 65.45 & 76.30 & 57.28 & 44.21 & 2.21 & 97.16\% \\
\sttm{} \venueyearlink{ICCV}{2025}{hyun2025sttm} & 65.52 & 82.37 & \underline{66.87} & \underline{77.54} & 58.14 & 44.94 & 2.29 & \underline{99.07\%} \\
\flashvid{} \venueyearlink{ICLR}{2026}{2026flashvid} & \underline{65.53} & \underline{82.83} & 66.72 & 77.52 & \underline{58.16} & \textbf{45.05} & 2.27 & 99.03\% \\
\learnpruner{} \venueyearlink{ICLR}{2026}{takezoe2026learnpruner} & 65.15 & 81.83 & 66.48 & 76.86 & 57.81 & 44.31 & \textbf{2.30} & 98.47\% \\
\cellcolor{oursburgundy}\textbf{FoveaTree (Ours)} & \cellcolor{oursburgundy}\textbf{65.67} & \cellcolor{oursburgundy}\textbf{83.73} & \cellcolor{oursburgundy}\textbf{67.42} & \cellcolor{oursburgundy}\textbf{78.04} & \cellcolor{oursburgundy}\textbf{58.59} & \cellcolor{oursburgundy}\underline{45.02} & \cellcolor{oursburgundy}\textbf{2.30} & \cellcolor{oursburgundy}\textbf{99.74\%} \\
\hline
\multicolumn{9}{c}{\textit{Retain 1,725 Tokens ($\downarrow$ \textbf{70.0\%})}} \\
\hline
\tome{} \venueyearlink{ICLR}{2023}{bolya2023tome} & 61.70 & 77.94 & 62.84 & 72.81 & 54.81 & 41.97 & 2.16 & 93.22\% \\
\fastv{} \venueyearlink{ECCV}{2024}{chen2024fastv} & 62.43 & 78.52 & 63.35 & 73.40 & 55.00 & 42.67 & 2.20 & 94.20\% \\
\visionzip{} \venueyearlink{CVPR}{2025}{yang2025visionzip} & 62.76 & 79.54 & 64.45 & 74.68 & 55.94 & 42.87 & 2.19 & 95.15\% \\
\framefusion{} \venueyearlink{ICCV}{2025}{fu2025framefusion} & 62.18 & 78.63 & 63.37 & 73.52 & 55.39 & 42.49 & 2.14 & 93.86\% \\
\sttm{} \venueyearlink{ICCV}{2025}{hyun2025sttm} & 64.54 & 81.90 & 66.07 & 76.75 & 57.47 & 44.13 & \textbf{2.29} & 98.04\% \\
\flashvid{} \venueyearlink{ICLR}{2026}{2026flashvid} & \underline{64.77} & \underline{82.20} & \underline{66.25} & \underline{76.92} & \underline{57.69} & \underline{44.44} & 2.27 & \underline{98.24\%} \\
\learnpruner{} \venueyearlink{ICLR}{2026}{takezoe2026learnpruner} & 63.86 & 80.54 & 65.08 & 75.80 & 56.71 & 43.60 & 2.22 & 96.50\% \\
\cellcolor{oursburgundy}\textbf{FoveaTree (Ours)} & \cellcolor{oursburgundy}\textbf{65.38} & \cellcolor{oursburgundy}\textbf{83.45} & \cellcolor{oursburgundy}\textbf{67.37} & \cellcolor{oursburgundy}\textbf{77.79} & \cellcolor{oursburgundy}\textbf{58.30} & \cellcolor{oursburgundy}\textbf{44.97} & \cellcolor{oursburgundy}\textbf{2.29} & \cellcolor{oursburgundy}\textbf{99.42\%} \\
\hline
\multicolumn{9}{c}{\textit{Retain 575 Tokens ($\downarrow$ \textbf{90.0\%})}} \\
\hline
\tome{} \venueyearlink{ICLR}{2023}{bolya2023tome} & 56.00 & 71.12 & 57.31 & 66.45 & 49.64 & 38.50 & 1.98 & 85.02\% \\
\fastv{} \venueyearlink{ECCV}{2024}{chen2024fastv} & 54.13 & 68.29 & 55.20 & 64.06 & 47.93 & 37.02 & 1.88 & 81.74\% \\
\visionzip{} \venueyearlink{CVPR}{2025}{yang2025visionzip} & 57.76 & 72.42 & 58.79 & 67.96 & 50.99 & 39.24 & 2.02 & 87.02\% \\
\framefusion{} \venueyearlink{ICCV}{2025}{fu2025framefusion} & 59.64 & 75.12 & 60.80 & 70.43 & 52.84 & 40.91 & 2.12 & 90.36\% \\
\sttm{} \venueyearlink{ICCV}{2025}{hyun2025sttm} & 63.38 & 80.18 & 64.95 & 75.00 & 56.04 & 43.54 & \underline{2.27} & 96.28\% \\
\flashvid{} \venueyearlink{ICLR}{2026}{2026flashvid} & \underline{63.95} & \underline{80.81} & \underline{65.68} & \underline{75.79} & \underline{56.73} & \underline{43.73} & 2.21 & \underline{96.67\%} \\
\learnpruner{} \venueyearlink{ICLR}{2026}{takezoe2026learnpruner} & 61.00 & 76.66 & 61.92 & 71.94 & 54.13 & 41.88 & 2.08 & 91.80\% \\
\cellcolor{oursburgundy}\textbf{FoveaTree (Ours)} & \cellcolor{oursburgundy}\textbf{64.92} & \cellcolor{oursburgundy}\textbf{82.58} & \cellcolor{oursburgundy}\textbf{67.03} & \cellcolor{oursburgundy}\textbf{77.45} & \cellcolor{oursburgundy}\textbf{57.97} & \cellcolor{oursburgundy}\textbf{44.69} & \cellcolor{oursburgundy}\textbf{2.28} & \cellcolor{oursburgundy}\textbf{98.81\%} \\
\hline
\end{tabular}%
}}
\end{minipage}\end{table*}

\subsection{Experimental Setup}

\paragraph{Benchmarks.}
We evaluate on six public video benchmarks spanning short- and long-video understanding, causal and temporal reasoning, and detailed captioning: Video-MME~\citep{fu2024videomme}, NExT-QA~\citep{xiao2021nextqa}, MVBench~\citep{li2024mvbench}, Valor32k~\citep{chen2023valor}, VNBench~\citep{zhao2025vnbench}, and VDC~\citep{chai2025auroracap}. Together, these tasks cover a broad and complementary set of multimodal understanding capabilities, including perception, causal and temporal reasoning, audio-visual association, long-video comprehension, and fine-grained generation. This ensures that pruning-induced degradation is measured across capability profiles rather than on any single skill. Benchmark details and evaluation protocols are provided in the Appendix.

\paragraph{Implementation Details.}
We evaluate InternVL3.5-8B and Qwen3-VL-8B with identical inputs and decoding settings, and compare ToMe~\citep{bolya2023tome}, FastV~\citep{chen2024fastv}, VisionZip~\citep{yang2025visionzip}, FrameFusion~\citep{fu2025framefusion}, STTM~\citep{hyun2025sttm}, FlashVID~\citep{2026flashvid}, and LearnPruner~\citep{takezoe2026learnpruner}. The proposed MGDE module uses layer-0 features, a root side length of 8, a per-frame leaf cap, and a merge-gain threshold of 0.005. The proposed ATP module is inserted at decoder layer 18. We run batch-one bfloat16 inference with SDPA and greedy decoding on one NVIDIA A800 80~GB GPU, synchronize CUDA timing, and exclude warm-up. Efficiency is measured on Video-MME with at most 64 frames. Exact retained-token counts and results on additional backbones are provided in the Appendix, together with full implementation details for all evaluated models, retention configurations, and benchmarks.

\setcounter{topnumber}{2}
\begin{table*}[!t]
\centering
\begin{minipage}[t]{\textwidth}
\centering
\caption{Efficiency analysis at 30\% and 10\% visual-token retention on InternVL3.5 on Video-MME.}
\label{tab:ten-percent-backbones-after-a}
{\scriptsize
\renewcommand{\arraystretch}{1.05}
\renewcommand{\multirowsetup}{\centering}
\setlength{\extrarowheight}{0.5pt}
\setlength{\tabcolsep}{0pt}
\resizebox{\linewidth}{!}{%
\begin{tabular}{>{\raggedright\arraybackslash}m{1.22in}|*{5}{>{\centering\arraybackslash}m{0.42695in}}|*{5}{>{\centering\arraybackslash}m{0.42695in}}}
\hline
\textbf{Method} & \textbf{Time} & \textbf{Mem} & \textbf{Latency} & \textbf{TFLOPs} & \textbf{Acc.} & \textbf{Time} & \textbf{Mem} & \textbf{Latency} & \textbf{TFLOPs} & \textbf{Acc.} \\
\hline
\rowcolor{gray!15}
\mbox{InternVL3.5 (16,384 Tokens)} & 111:47 & 20.26 & 2.484 & 322.0 & 65.67 & 111:47 & 20.26 & 2.484 & 322.0 & 65.67 \\
\hline
& \multicolumn{5}{c|}{\color{black}\textit{Retain 4,928 Tokens ($\downarrow$ \textbf{70.0\%})}} & \multicolumn{5}{c}{\color{black}\textit{Retain 1,664 Tokens ($\downarrow$ \textbf{90.0\%})}} \\
\hline
\rowcolor{gray!15}
\cellcolor{white}\tome{} \venueyearlink{ICLR}{2023}{bolya2023tome} & \textcolor{black!40}{74:58} & \textcolor{black!40}{17.67} & \textcolor{black!40}{1.666} & \textcolor{black!40}{86.5} & 61.59 & \textcolor{black!40}{89:28} & \textcolor{black!40}{16.90} & \textcolor{black!40}{1.988} & \textcolor{black!40}{33.6} & 56.92 \\
\rowcolor{gray!15}
\cellcolor{white}\fastv{} \venueyearlink{ECCV}{2024}{chen2024fastv} & \textcolor{black!40}{106:34} & \textcolor{black!40}{19.09} & \textcolor{black!40}{2.368} & \textcolor{black!40}{204.2} & 61.22 & \textcolor{black!40}{98:06} & \textcolor{black!40}{19.09} & \textcolor{black!40}{2.180} & \textcolor{black!40}{177.8} & 58.22 \\
\rowcolor{gray!15}
\cellcolor{white}\visionzip{} \venueyearlink{CVPR}{2025}{yang2025visionzip} & \textcolor{black!40}{49:11} & \textcolor{black!40}{17.67} & \textcolor{black!40}{1.093} & \textcolor{black!40}{86.5} & 62.56 & \textcolor{black!40}{44:25} & \textcolor{black!40}{16.90} & \textcolor{black!40}{0.987} & \textcolor{black!40}{33.6} & 60.10 \\
\rowcolor{gray!15}
\cellcolor{white}\framefusion{} \venueyearlink{ICCV}{2025}{fu2025framefusion} & \textcolor{black!40}{54:22} & \textcolor{black!40}{17.94} & \textcolor{black!40}{1.208} & \textcolor{black!40}{93.0} & 62.24 & \textcolor{black!40}{35:52} & \textcolor{black!40}{17.94} & \textcolor{black!40}{0.797} & \textcolor{black!40}{43.9} & 60.42 \\
\rowcolor{gray!15}
\cellcolor{white}\sttm{} \venueyearlink{ICCV}{2025}{hyun2025sttm} & \textcolor{black!40}{82:53} & \textcolor{black!40}{19.34} & \textcolor{black!40}{1.842} & \textcolor{black!40}{204.2} & 64.33 & \textcolor{black!40}{86:02} & \textcolor{black!40}{19.15} & \textcolor{black!40}{1.912} & \textcolor{black!40}{177.8} & 63.58 \\
\rowcolor{oursburgundy}
\cellcolor{white}\flashvid{} \venueyearlink{ICLR}{2026}{2026flashvid} & 95:16 & \textbf{17.95} & 2.117 & 197.3 & \underline{65.12} & \underline{66:12} & \textbf{17.00} & \underline{1.471} & 136.7 & \underline{63.96} \\
\rowcolor{oursburgundy}
\cellcolor{white}\learnpruner{} \venueyearlink{ICLR}{2026}{takezoe2026learnpruner} & \underline{92:10} & 19.74 & \underline{2.048} & \textbf{143.2} & 63.66 & 69:56 & \underline{18.92} & 1.554 & \underline{132.9} & 62.87 \\
\rowcolor{oursburgundy}
\cellcolor{white}\textbf{FoveaTree (Ours)} & \textbf{90:16} & \underline{19.36} & \textbf{2.006} & \underline{150.8} & \textbf{66.01} & \textbf{65:07} & 19.36 & \textbf{1.447} & \textbf{125.1} & \textbf{65.41} \\
\hline
\end{tabular}%
}}
\end{minipage}
\end{table*}

\begin{figure*}[!t]
    \centering
    \includegraphics[width=\linewidth]{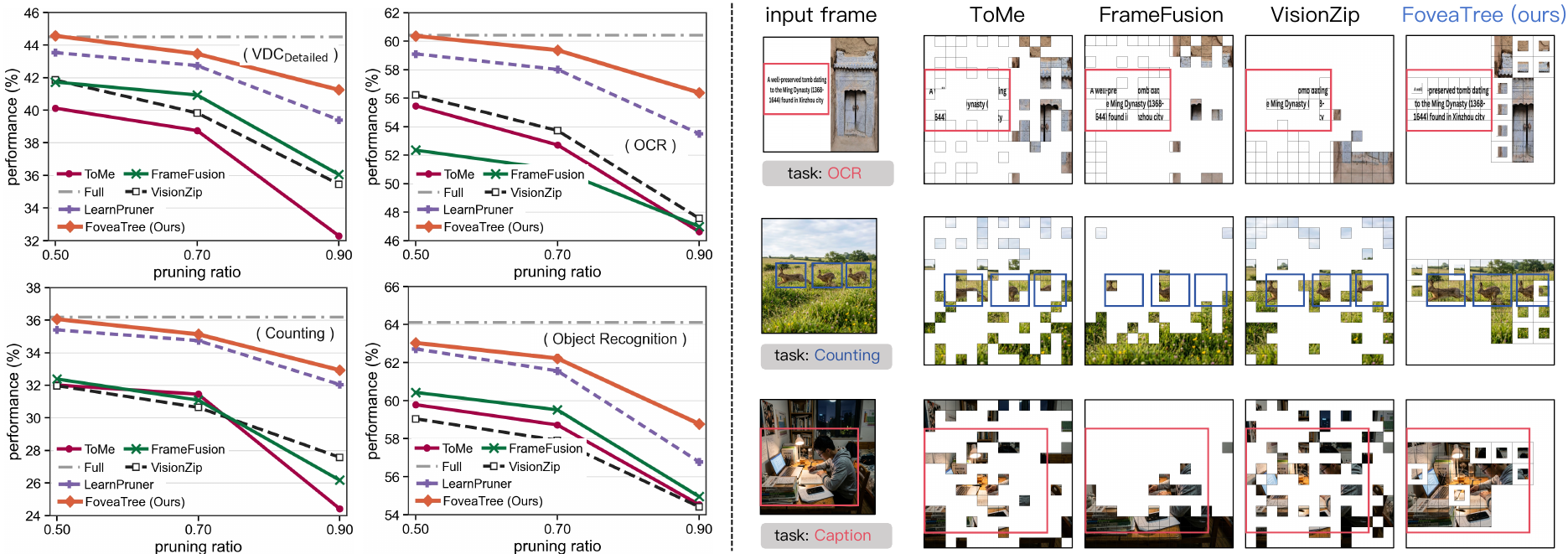}
    \caption{Fine-grained performance and retained-token visualizations. FoveaTree achieves the strongest results across the shown token-retention ratios, while the right panel illustrates that it preserves relevant regions more completely than the compared methods under matched budgets.}
    \label{fig:exp-hb}
\end{figure*}

\subsection{Comparison with State-of-the-Art Methods}

Tables~\ref{tab:main-backbone-a} and~\ref{tab:main-backbone-b} report all seven metrics at 50\%, 30\%, and 10\% visual-token retention on both evaluated backbones. The five Video QA benchmarks use answer accuracy, while VDC reports Acc. and Score. Avg. (\%) equally averages each reported metric relative to its corresponding Full result. FoveaTree consistently achieves the highest Avg. for every reported setting on both backbones, and the margin over the strongest competing baseline widens as the token budget decreases.

\paragraph{Results on InternVL3.5-8B.}
On InternVL3.5-8B, FoveaTree obtains 100.62\%, 100.69\%, and 99.37\% Avg. at 30\%, 50\%, and 10\% retention, respectively. At 30\%, it leads on five of the seven metrics, including 66.01 on Video-MME and 41.39/1.95 on VDC$_{\mathrm{Acc.}}$/VDC$_{\mathrm{Score}}$, and ranks a close second to the strongest baseline on the two remaining metrics of NExT-QA and VNBench.

\paragraph{Results on Qwen3-VL-8B.}
On Qwen3-VL-8B, FoveaTree obtains 99.42\%, 99.74\%, and 98.81\% Avg. at 30\%, 50\%, and 10\% retention, respectively, and leads on all seven metrics at 30\%, including 65.38 on Video-MME and 44.97/2.29 on VDC$_{\mathrm{Acc.}}$/VDC$_{\mathrm{Score}}$. Across both backbones, FoveaTree remains ahead of the compared methods at every budget, and its gains extend to VDC detailed captioning, which requires accurate descriptions of objects, events, and visual details across frames.

\subsection{Efficiency Analysis}

To measure the efficiency of FoveaTree, we record total inference time, peak memory, latency, and TFLOPs on Video-MME together with Video-MME accuracy from the complete benchmark evaluation, as reported in Table~\ref{tab:ten-percent-backbones-after-a}. At 30\% retention on InternVL3.5-8B, FoveaTree reduces TFLOPs by 53.2\% relative to Full (322.0 to 150.8) with a $1.24\times$ latency speedup (2.484~s to 2.006~s), while peak memory stays within 4.4\% of the unpruned model (20.26 to 19.36~GiB). Relative to the earlier pruning methods (shown in gray), FoveaTree leads in Video-MME accuracy by a large margin; relative to the more recent and stronger baselines, it attains the lowest inference time at both budgets while remaining the most accurate, demonstrating the superior accuracy-efficiency trade-off.

\setcounter{topnumber}{3}

\begin{table*}[t]
\centering
\begin{minipage}[t]{3.32in}
\centering
\makebox[\linewidth][l]{%
\begin{minipage}[t]{2.868in}
\centering
\captionof{table}{Component ablation of MGDE and ATP.}
\label{tab:component-ablation}
\end{minipage}%
}
{\tablebodyformat
\setlength{\tabcolsep}{0pt}
\makebox[\linewidth][l]{%
\begin{tabular}{>{\centering\arraybackslash}m{0.35in}@{}>{\centering\arraybackslash}m{0.35in}|>{\centering\arraybackslash}m{0.542in}>{\centering\arraybackslash}m{0.542in}|>{\centering\arraybackslash}m{0.542in}>{\centering\arraybackslash}m{0.542in}}
\hline
\textbf{MGDE} & \textbf{ATP} & \textbf{VideoMME} & \textbf{VDC$_{\mathrm{Acc.}}$} & \textbf{Avg.} & \textbf{Lat.} \\
\hline
\rowcolor{gray!15}
$\times$ & $\times$ & 65.67 & 40.47 & 100.00\% & 2.484 \\
\checkmark & $\times$ & \underline{65.50} & \underline{41.08} & \underline{100.62\%} & 2.247 \\
$\times$ & \checkmark & 65.44 & 41.00 & 100.48\% & \underline{2.086} \\
\rowcolor{oursburgundy}\cellcolor{oursburgundy}\checkmark & \cellcolor{oursburgundy}\checkmark & \cellcolor{oursburgundy}\textbf{66.01} & \cellcolor{oursburgundy}\textbf{41.39} & \cellcolor{oursburgundy}\textbf{101.40\%} & \cellcolor{oursburgundy}\textbf{2.006} \\
\hline
\end{tabular}%
}}
\end{minipage}\hfill
\begin{minipage}[t]{2.08in}
\centering
\captionof{table}{MGDE feature-layer study.}
\label{tab:mgde-layer}
{\tablebodyformat
\setlength{\tabcolsep}{0pt}
\makebox[\linewidth][r]{%
\begin{tabular}{>{\centering\arraybackslash}m{0.45in}|>{\centering\arraybackslash}m{0.542in}>{\centering\arraybackslash}m{0.542in}|>{\centering\arraybackslash}m{0.542in}}
\hline
\textbf{Layer} & \textbf{VideoMME} & \textbf{VDC$_{\mathrm{Acc.}}$} & \textbf{Avg.} \\
\hline
\rowcolor{oursburgundy}\cellcolor{oursburgundy}0  & \cellcolor{oursburgundy}\textbf{66.01} & \cellcolor{oursburgundy}\textbf{41.39} & \cellcolor{oursburgundy}\textbf{101.40\%} \\
12 & \underline{65.48} & \underline{40.99} & \underline{100.50\%} \\
24 & 65.18 & 40.78 & 100.01\% \\
32 & 62.55 & 38.65 & 95.37\% \\
\hline
\end{tabular}%
}}
\end{minipage}

\par\vspace{6pt}
\begin{minipage}[t]{2.08in}
\centering
\captionof{table}{ATP insertion-layer study.}
\label{tab:atp-layer}
{\tablebodyformat
\setlength{\tabcolsep}{0pt}
\makebox[\linewidth][r]{%
\begin{tabular}{>{\centering\arraybackslash}m{0.45in}|>{\centering\arraybackslash}m{0.542in}>{\centering\arraybackslash}m{0.542in}|>{\centering\arraybackslash}m{0.542in}}
\hline
\textbf{Layer} & \textbf{VideoMME} & \textbf{VDC$_{\mathrm{Acc.}}$} & \textbf{Avg.} \\
\hline
6 & 65.29 & 40.84 & 100.17\% \\
12 & 65.43 & 40.97 & 100.43\% \\
\rowcolor{oursburgundy}\cellcolor{oursburgundy}18 & \cellcolor{oursburgundy}\textbf{66.01} & \cellcolor{oursburgundy}\textbf{41.39} & \cellcolor{oursburgundy}\textbf{101.40\%} \\
24 & \underline{65.88} & \underline{41.27} & \underline{101.15\%} \\
\hline
\end{tabular}%
}}
\end{minipage}\hfill
\begin{minipage}[t]{3.32in}
\centering
\captionof{table}{ATP scoring and selection strategies.}
\label{tab:internal-ablation}
{\fontsize{6.856pt}{8.771pt}\selectfont
\renewcommand{\arraystretch}{1.0}
\renewcommand{\multirowsetup}{\centering}
\setlength{\extrarowheight}{0.452pt}
\setlength{\tabcolsep}{0pt}
\makebox[\linewidth][r]{%
\resizebox{3.1465in}{!}{%
\begin{tabular}{>{\centering\arraybackslash}m{0.82in}|>{\centering\arraybackslash}m{0.50in}>{\centering\arraybackslash}m{0.50in}|>{\centering\arraybackslash}m{0.50in}>{\centering\arraybackslash}m{0.50in}}
\hline
\textbf{Strategy} & \textbf{VideoMME} & \textbf{VDC$_{\mathrm{Acc.}}$} & \textbf{Avg.} & \textbf{Lat.} \\
\hline
Flat Top-$k$ & 65.05 & 40.62 & 99.71\% & \textbf{1.896} \\
Debiased Top-$k$ & 65.39 & 40.93 & 100.36\% & 2.700 \\
Debiased ATP & \textbf{66.11} & \underline{41.39} & \textbf{101.47\%} & 2.813 \\
\rowcolor{oursburgundy}\cellcolor{oursburgundy}ATP (Ours) & \cellcolor{oursburgundy}\underline{66.01} & \cellcolor{oursburgundy}\textbf{41.39} & \cellcolor{oursburgundy}\underline{101.40\%} & \cellcolor{oursburgundy}\underline{2.006} \\
\hline
\end{tabular}%
}}}
\end{minipage}
\par\vspace{9pt}
\begin{minipage}{\linewidth}
\centering
    \includegraphics[width=\linewidth]{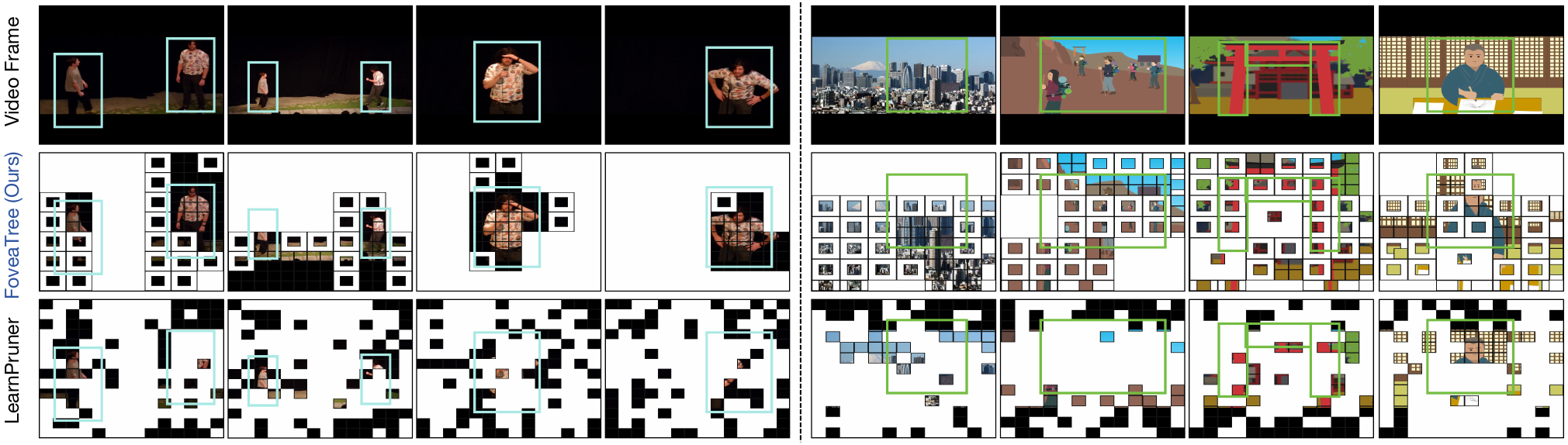}
    \captionof{figure}{Qualitative comparison of LearnPruner and FoveaTree with video frames. The examples illustrate that FoveaTree preserves key objects and their surrounding regions more completely.}
    \label{fig:exp-final}
\end{minipage}
\end{table*}

\setcounter{topnumber}{3}

\subsection{Ablation Study}

We conduct ablations on Video-MME and VDC using InternVL3.5-8B. Configurations containing ATP use a nominal 30\% per-frame target, whereas MGDE-only reports its adaptive output before ATP budget refinement. Avg. equally weights the two reported metrics after each is normalized by the corresponding Full result of the unpruned InternVL3.5-8B model on the very same inputs.

\paragraph{Component effectiveness.}
Table~\ref{tab:component-ablation} evaluates MGDE and ATP while recording the latency of each configuration. The first row retains the original visual sequence. MGDE-only constructs the quadtree and performs density-guided token operations before prefill without ATP. ATP-only applies query-guided pruning to a complete per-frame quadtree whose leaves are the original projected tokens. Combining both stages reaches 101.40\% Avg. under the nominal 30\% target, improving Avg. by 0.78 and 0.92 points over MGDE-only and ATP-only while reducing latency to 2.006~s and preserving complementary regional and query-guided gains across both stages.

\paragraph{Layer selection.}
Under the nominal 30\% target, we study the feature and decoder layers used by both stages. Table~\ref{tab:mgde-layer} shows that layer-0 vision features give MGDE the best encoder-layer result, reaching 101.40\% Avg. Table~\ref{tab:atp-layer} shows that ATP at decoder layer 18 gives the highest Avg. among the tested layers, also reaching 101.40\%. We use layers 0 and 18 for regional analysis and for query-guided pruning in all subsequent experiments and qualitative visualizations.

\paragraph{ATP strategy.}
Table~\ref{tab:internal-ablation} compares ATP with flat Top-$k$ and debiased variants at 30\% retention, reporting both Avg. and latency. Following representation-difference calibration~\citep{liu2024vti,zhao2025attentiondebiasing}, debiasing subtracts layer-18 hidden states obtained from an all-zero reference video under the same query before recomputing text-to-vision attention, thereby reducing video-independent bias. ATP reaches 101.40\% Avg. and 2.006~s, within 0.07 points of Debiased ATP while avoiding its additional reference pass at 2.813~s, providing direct evidence of stronger robustness to Attention Drift at substantially lower latency than every debiased alternative in this comparison.

\subsection{Visualization}
\label{sec:visualization}

Figs.~\ref{fig:exp-hb} and~\ref{fig:exp-final} visualize FoveaTree with respect to the two challenges introduced in the Introduction. Fig.~\ref{fig:exp-hb} reports fine-grained performance across retention ratios, where FoveaTree consistently outperforms similarity-based methods, and its right panel shows more complete preservation of regional evidence, supporting hierarchy-aware pruning. Fig.~\ref{fig:exp-final} presents two Video-MME examples in which, under the same token budget, FoveaTree is less affected by Attention Drift than LearnPruner and more accurately preserves query-relevant core regions and their surrounding evidence intact.

\section{Conclusion}

We presented FoveaTree, a training-free framework motivated by Hierarchy-blind Pruning and Attention Drift in Video LLMs. MGDE reduces redundant visual units through a compressibility-derived spatial hierarchy, while ATP propagates query-conditioned attention through the surviving tree. Experiments on two Video LLMs and six benchmarks at three retention budgets demonstrate stronger performance retention with competitive efficiency, establishing regional hierarchy as a practical bridge between query-independent compression and query-aware selection. The framework is more complex than basic pruning methods and lacks explicit temporal processing; future work will simplify the pipeline and extend the hierarchy across time to model motion and event dynamics.

\section*{Ethics Statement}

This work does not involve human participants, human subjects, personally identifiable information, or the collection of new personal data. We evaluate on publicly released video benchmarks, including Video-MME, NExT-QA, MVBench, Valor32k, VNBench, and VDC, following their intended evaluation protocols and applicable usage terms. The proposed method is an inference-time visual-token pruning framework and does not generate new training data or modify the parameters of the evaluated Video LLMs. As with other video-language systems, the evaluated models may inherit biases, hallucinations, and errors from their underlying data and pretrained models. Token pruning may additionally remove visual evidence in difficult cases, particularly for fine-grained, temporal, or safety-critical content. We therefore view FoveaTree as an efficiency technique rather than a substitute for human oversight in high-stakes applications.

\section*{Reproducibility Statement}

We evaluate FoveaTree on four publicly available Video LLM backbones, six public video benchmarks, and three visual-token retention budgets. The paper and supplementary material describe the MGDE and ATP algorithms, regional distortion and attention-score formulations, quadtree construction, budget resolution procedure, model configurations, benchmark protocols, decoding settings, hardware, timing methodology, and token-budget audits. The supplementary material also provides additional backbone results, component and layer ablations, efficiency analyses, and implementation details for the compared baselines. Our code, configuration files, and evaluation scripts will be released in the final version of the paper to ensure that all reported results can be reproduced.

\section*{AI Use Statement}

Generative AI tools assisted with translation, manuscript drafting, literature search, reference formatting, LaTeX typesetting, and plotting code. They provided feedback on methodological descriptions and experimental analyses, assisted with mathematical derivations and their exposition, and supported numerical calculations when preparing results. The authors retain responsibility for the final manuscript, including all claims, references, numerical results, and AI-assisted artifacts.

\begingroup
\setlength{\bibsep}{3.95pt plus 1pt minus 0.5pt}
\bibliographystyle{iclr2027_conference}
\bibliography{appendix_references}
\endgroup

\clearpage
\setcounter{topnumber}{2}
\setlength{\intextsep}{12pt plus 2pt minus 2pt}
\addtocontents{toc}{\protect\setcounter{tocdepth}{2}}
\begin{center}
{\LARGE\bfseries Supplementary Material}
\end{center}
\vspace{6pt}
\appendix
\renewcommand{\contentsname}{CONTENTS}
\tableofcontents
\newpage
\begin{center}
\begin{minipage}{\textwidth}
\centering
\captionof{table}{Qwen2.5-VL-7B results at 10\%, 30\%, and 50\% visual-token retention. Bold and underlined values mark the best and second-best compressed results; Full provides the unpruned reference.}
\label{tab:app-qwen25-results}
{\scriptsize
\renewcommand{\arraystretch}{1.05}
\renewcommand{\multirowsetup}{\centering}
\setlength{\extrarowheight}{0.5pt}
\setlength{\tabcolsep}{0pt}
\resizebox{\linewidth}{!}{%
\begin{tabular}{>{\raggedright\arraybackslash}m{1.22in}|*{5}{>{\centering\arraybackslash}m{0.5435in}}|*{2}{>{\centering\arraybackslash}m{0.5435in}}|>{\centering\arraybackslash}m{0.465in}}
\hline
\multirow[c]{2}{1.2in}{\raggedright\textbf{Method}} & \multicolumn{5}{c|}{\textbf{Video QA}} & \multicolumn{2}{c|}{\textbf{Video Caption}} & \multirow[c]{2}{0.465in}{\centering\textbf{Avg.}} \\
\cline{2-8}
& \textbf{VideoMME} & \textbf{NExT-QA} & \textbf{MVBench} & \textbf{Valor32k} & \textbf{VNBench} & \textbf{VDC$_{\mathrm{Acc.}}$} & \textbf{VDC$_{\mathrm{Score}}$} & \\
\hline
\rowcolor{gray!15}
\mbox{Qwen2.5-VL (5,753 Tokens)} & 61.85 & 82.32 & 69.16 & 76.94 & 66.32 & 45.40 & 2.34 & 100.00\% \\
\hline
\multicolumn{9}{c}{\textit{Retain 2,876 Tokens ($\downarrow$ \textbf{50.0\%})}} \\
\hline
\tome{} \venueyearlink{ICLR}{2023}{bolya2023tome} & 58.80 & 78.45 & 65.73 & 73.27 & 63.18 & 43.44 & 2.23 & 95.27\% \\
\fastv{} \venueyearlink{ECCV}{2024}{chen2024fastv} & 59.64 & 79.42 & 66.87 & 74.26 & 64.16 & 43.70 & 2.26 & 96.53\% \\
\visionzip{} \venueyearlink{CVPR}{2025}{yang2025visionzip} & 60.19 & 79.95 & 67.14 & 74.90 & 64.65 & 44.28 & 2.27 & 97.27\% \\
\framefusion{} \venueyearlink{ICCV}{2025}{fu2025framefusion} & 59.76 & 79.47 & 66.95 & 74.31 & 64.05 & 43.78 & 2.26 & 96.59\% \\
\sttm{} \venueyearlink{ICCV}{2025}{hyun2025sttm} & \underline{61.35} & 81.54 & \textbf{68.74} & 76.29 & \underline{65.90} & \underline{44.97} & \underline{2.32} & \underline{99.19\%} \\
\flashvid{} \venueyearlink{ICLR}{2026}{2026flashvid} & \textbf{61.39} & \underline{81.63} & \underline{68.64} & \underline{76.34} & 65.50 & 44.95 & 2.32 & 99.11\% \\
\learnpruner{} \venueyearlink{ICLR}{2026}{takezoe2026learnpruner} & 60.46 & 80.37 & 67.80 & 75.17 & 64.85 & 44.48 & 2.29 & 97.82\% \\
\cellcolor{oursburgundy}\textbf{FoveaTree (Ours)} & \cellcolor{oursburgundy}61.30 & \cellcolor{oursburgundy}\textbf{82.16} & \cellcolor{oursburgundy}68.55 & \cellcolor{oursburgundy}\textbf{76.66} & \cellcolor{oursburgundy}\textbf{65.91} & \cellcolor{oursburgundy}\textbf{45.17} & \cellcolor{oursburgundy}\textbf{2.33} & \cellcolor{oursburgundy}\textbf{99.45\%} \\
\hline
\multicolumn{9}{c}{\textit{Retain 1,725 Tokens ($\downarrow$ \textbf{70.0\%})}} \\
\hline
\tome{} \venueyearlink{ICLR}{2023}{bolya2023tome} & 56.33 & 74.92 & 63.21 & 70.23 & 60.60 & 41.41 & 2.13 & 91.20\% \\
\fastv{} \venueyearlink{ECCV}{2024}{chen2024fastv} & 56.74 & 75.63 & 63.45 & 70.71 & 61.07 & 41.78 & 2.15 & 91.89\% \\
\visionzip{} \venueyearlink{CVPR}{2025}{yang2025visionzip} & 58.28 & 77.60 & 65.40 & 72.74 & 62.60 & 42.70 & 2.21 & 94.36\% \\
\framefusion{} \venueyearlink{ICCV}{2025}{fu2025framefusion} & 57.53 & 76.87 & 64.59 & 71.81 & 61.89 & 42.50 & 2.18 & 93.32\% \\
\sttm{} \venueyearlink{ICCV}{2025}{hyun2025sttm} & 60.81 & 81.08 & 67.96 & 75.54 & 65.21 & 44.54 & 2.30 & 98.28\% \\
\flashvid{} \venueyearlink{ICLR}{2026}{2026flashvid} & \underline{60.99} & \underline{81.50} & \underline{68.31} & \underline{76.16} & \underline{65.63} & \textbf{44.77} & \textbf{2.31} & \underline{98.81\%} \\
\learnpruner{} \venueyearlink{ICLR}{2026}{takezoe2026learnpruner} & 59.52 & 79.35 & 66.51 & 74.22 & 63.80 & 43.70 & 2.25 & 96.27\% \\
\cellcolor{oursburgundy}\textbf{FoveaTree (Ours)} & \cellcolor{oursburgundy}\textbf{61.44} & \cellcolor{oursburgundy}\textbf{82.23} & \cellcolor{oursburgundy}\textbf{68.58} & \cellcolor{oursburgundy}\textbf{76.64} & \cellcolor{oursburgundy}\textbf{65.85} & \cellcolor{oursburgundy}\underline{44.69} & \cellcolor{oursburgundy}\underline{2.31} & \cellcolor{oursburgundy}\textbf{99.21\%} \\
\hline
\multicolumn{9}{c}{\textit{Retain 575 Tokens ($\downarrow$ \textbf{90.0\%})}} \\
\hline
\tome{} \venueyearlink{ICLR}{2023}{bolya2023tome} & 51.43 & 68.34 & 57.62 & 64.09 & 55.16 & 37.67 & 1.94 & 83.12\% \\
\fastv{} \venueyearlink{ECCV}{2024}{chen2024fastv} & 51.09 & 67.66 & 56.96 & 63.30 & 54.66 & 37.52 & 1.93 & 82.42\% \\
\visionzip{} \venueyearlink{CVPR}{2025}{yang2025visionzip} & 57.18 & 75.97 & 63.90 & 71.15 & 61.36 & 41.76 & 2.16 & 92.35\% \\
\framefusion{} \venueyearlink{ICCV}{2025}{fu2025framefusion} & 55.72 & 73.99 & 62.21 & 69.30 & 59.75 & 40.78 & 2.11 & 90.01\% \\
\sttm{} \venueyearlink{ICCV}{2025}{hyun2025sttm} & 60.04 & 79.93 & 67.37 & 74.63 & 64.37 & \underline{43.97} & \underline{2.27} & 97.07\% \\
\flashvid{} \venueyearlink{ICLR}{2026}{2026flashvid} & \underline{60.61} & \underline{80.79} & \underline{67.65} & \underline{75.38} & \underline{65.04} & \textbf{44.43} & \textbf{2.29} & \underline{97.96\%} \\
\learnpruner{} \venueyearlink{ICLR}{2026}{takezoe2026learnpruner} & 58.11 & 77.16 & 64.67 & 71.94 & 62.34 & 42.45 & 2.19 & 93.68\% \\
\cellcolor{oursburgundy}\textbf{FoveaTree (Ours)} & \cellcolor{oursburgundy}\textbf{61.37} & \cellcolor{oursburgundy}\textbf{82.02} & \cellcolor{oursburgundy}\textbf{68.76} & \cellcolor{oursburgundy}\textbf{76.57} & \cellcolor{oursburgundy}\textbf{65.98} & \cellcolor{oursburgundy}43.40 & \cellcolor{oursburgundy}2.24 & \cellcolor{oursburgundy}\textbf{98.37\%} \\
\hline
\end{tabular}%
}}
\end{minipage}
\end{center}

\section{More Experimental Results}
\label{app:analysis}

\subsection{Additional Experiments on Qwen2.5-VL}
\label{app:qwen25-results}

The main paper evaluates InternVL3.5-8B and Qwen3-VL-8B. To examine generality beyond these two backbones, this subsection applies the same evaluation protocol to Qwen2.5-VL-7B. Table~\ref{tab:app-qwen25-results} reports the full seven-metric protocol at 50\%, 30\%, and 10\% retention budgets, following the identical frame sampling and decoding configuration. These results quantify how each pruning method behaves across the QA benchmarks and the captioning score under each budget.
Qwen2.5-VL-7B~\citep{bai2025qwen25vl} uses a native dynamic-resolution vision transformer with window attention and 2D-RoPE. Its spatial token count varies with input resolution, and MGDE and ATP operate on each temporal unit's native grid.

\subsection{Additional Experiments on LLaVA-OneVision}
\label{app:llava-ov-results}

To examine generality beyond the two backbones used in the main comparison, we apply the same evaluation protocol to LLaVA-OneVision-7B~\citep{li2025llavaonevision}. Table~\ref{tab:app-llava-ov-results} reports the complete seven-metric protocol at 50\%, 30\%, and 10\% retention. Frame sampling follows the model's official video configuration, and all methods share identical inputs and decoding settings. FoveaTree maintains competitive performance across all three retention budgets, with particularly strong results on the Video QA benchmarks at 10\% retention.
Unlike the two main backbones, LLaVA-OneVision-7B pairs a SigLIP-SO400M vision encoder with a Qwen2-7B decoder and encodes each sampled frame (one native temporal unit) into a fixed low-resolution grid of 196 tokens, so the visual sequence is uniform across units rather than resolution-adaptive. MGDE uses the same quadtree construction rule on each unit, and ATP is inserted at decoder layer 18, transferring the protocol without architectural modification.

\begin{table*}[!t]
\centering
\begin{minipage}[t]{\textwidth}
\centering
\caption{LLaVA-OV-7B results at 10\%, 30\%, and 50\% visual-token retention. Bold and underlined values mark the best and second-best compressed results; Full provides the unpruned reference.}
\label{tab:app-llava-ov-results}
{\scriptsize
\renewcommand{\arraystretch}{1.05}
\renewcommand{\multirowsetup}{\centering}
\setlength{\extrarowheight}{0.5pt}
\setlength{\tabcolsep}{0pt}
\resizebox{\linewidth}{!}{%
\begin{tabular}{>{\raggedright\arraybackslash}m{1.22in}|*{5}{>{\centering\arraybackslash}m{0.5435in}}|*{2}{>{\centering\arraybackslash}m{0.5435in}}|>{\centering\arraybackslash}m{0.465in}}
\hline
\multirow[c]{2}{1.2in}{\raggedright\textbf{Method}} & \multicolumn{5}{c|}{\textbf{Video QA}} & \multicolumn{2}{c|}{\textbf{Video Caption}} & \multirow[c]{2}{0.465in}{\centering\textbf{Avg.}} \\
\cline{2-8}
& \textbf{VideoMME} & \textbf{NExT-QA} & \textbf{MVBench} & \textbf{Valor32k} & \textbf{VNBench} & \textbf{VDC$_{\mathrm{Acc.}}$} & \textbf{VDC$_{\mathrm{Score}}$} & \\
\hline
\rowcolor{gray!15}
\mbox{LLaVA-OV (12,544 Tokens)} & 58.56 & 80.57 & 58.26 & 74.11 & 62.87 & 41.85 & 2.17 & 100.00\% \\
\hline
\multicolumn{9}{c}{\textit{Retain 6,272 Tokens ($\downarrow$ \textbf{50.0\%})}} \\
\hline
\tome{} \venueyearlink{ICLR}{2023}{bolya2023tome} & 57.17 & 78.86 & 56.84 & 72.48 & 61.51 & 41.13 & 2.12 & 97.81\% \\
\fastv{} \venueyearlink{ECCV}{2024}{chen2024fastv} & 55.92 & 76.98 & 55.79 & 70.83 & 60.24 & 39.88 & 2.08 & 95.62\% \\
\visionzip{} \venueyearlink{CVPR}{2025}{yang2025visionzip} & 57.04 & 78.32 & 56.59 & 72.21 & 61.35 & 40.86 & 2.11 & 97.38\% \\
\framefusion{} \venueyearlink{ICCV}{2025}{fu2025framefusion} & 56.89 & 78.21 & 56.72 & 71.97 & 61.05 & 40.57 & 2.11 & 97.14\% \\
\sttm{} \venueyearlink{ICCV}{2025}{hyun2025sttm} & 58.07 & 79.79 & 57.91 & 73.47 & 62.46 & 41.44 & 2.15 & 99.17\% \\
\flashvid{} \venueyearlink{ICLR}{2026}{2026flashvid} & \textbf{58.71} & \underline{80.76} & \underline{58.38} & \underline{74.31} & \underline{62.66} & \underline{41.85} & \textbf{2.17} & \underline{100.09\%} \\
\learnpruner{} \venueyearlink{ICLR}{2026}{takezoe2026learnpruner} & 57.81 & 79.43 & 57.70 & 73.11 & 62.08 & 41.41 & 2.14 & 98.76\% \\
\cellcolor{oursburgundy}\textbf{FoveaTree (Ours)} & \cellcolor{oursburgundy}\underline{58.56} & \cellcolor{oursburgundy}\textbf{81.46} & \cellcolor{oursburgundy}\textbf{58.90} & \cellcolor{oursburgundy}\textbf{74.93} & \cellcolor{oursburgundy}\textbf{63.56} & \cellcolor{oursburgundy}\textbf{42.31} & \cellcolor{oursburgundy}\underline{2.16} & \cellcolor{oursburgundy}\textbf{100.72\%} \\
\hline
\multicolumn{9}{c}{\textit{Retain 3,776 Tokens ($\downarrow$ \textbf{70.0\%})}} \\
\hline
\tome{} \venueyearlink{ICLR}{2023}{bolya2023tome} & 55.13 & 75.81 & 55.06 & 69.93 & 59.39 & 39.47 & 2.05 & 94.34\% \\
\fastv{} \venueyearlink{ECCV}{2024}{chen2024fastv} & 54.17 & 74.66 & 53.88 & 68.70 & 58.40 & 38.84 & 2.01 & 92.67\% \\
\visionzip{} \venueyearlink{CVPR}{2025}{yang2025visionzip} & 56.52 & 77.80 & 56.45 & 71.78 & 60.79 & 40.32 & 2.10 & 96.66\% \\
\framefusion{} \venueyearlink{ICCV}{2025}{fu2025framefusion} & 56.44 & 77.96 & 56.39 & 71.67 & 60.80 & 40.60 & 2.10 & 96.73\% \\
\sttm{} \venueyearlink{ICCV}{2025}{hyun2025sttm} & 57.49 & 79.25 & 57.16 & 72.65 & 61.73 & 40.99 & 2.13 & 98.14\% \\
\flashvid{} \venueyearlink{ICLR}{2026}{2026flashvid} & \underline{58.37} & \underline{80.74} & \underline{58.12} & \underline{74.21} & \underline{62.86} & \underline{41.71} & \textbf{2.17} & \underline{99.92\%} \\
\learnpruner{} \venueyearlink{ICLR}{2026}{takezoe2026learnpruner} & 56.51 & 77.89 & 56.17 & 71.70 & 60.66 & 40.40 & 2.10 & 96.59\% \\
\cellcolor{oursburgundy}\textbf{FoveaTree (Ours)} & \cellcolor{oursburgundy}\textbf{58.46} & \cellcolor{oursburgundy}\textbf{81.28} & \cellcolor{oursburgundy}\textbf{58.76} & \cellcolor{oursburgundy}\textbf{74.75} & \cellcolor{oursburgundy}\textbf{63.41} & \cellcolor{oursburgundy}\textbf{42.21} & \cellcolor{oursburgundy}\underline{2.15} & \cellcolor{oursburgundy}\textbf{100.46\%} \\
\hline
\multicolumn{9}{c}{\textit{Retain 1,280 Tokens ($\downarrow$ \textbf{90.0\%})}} \\
\hline
\tome{} \venueyearlink{ICLR}{2023}{bolya2023tome} & 49.43 & 67.90 & 49.30 & 62.67 & 53.08 & 35.25 & 1.83 & 84.41\% \\
\fastv{} \venueyearlink{ECCV}{2024}{chen2024fastv} & 47.96 & 65.64 & 47.57 & 60.44 & 51.36 & 34.30 & 1.77 & 81.68\% \\
\visionzip{} \venueyearlink{CVPR}{2025}{yang2025visionzip} & 52.10 & 71.53 & 51.80 & 65.94 & 55.97 & 37.02 & 1.93 & 88.87\% \\
\framefusion{} \venueyearlink{ICCV}{2025}{fu2025framefusion} & 51.10 & 70.14 & 50.75 & 64.66 & 54.87 & 36.40 & 1.89 & 87.15\% \\
\sttm{} \venueyearlink{ICCV}{2025}{hyun2025sttm} & 56.54 & 77.82 & 56.47 & 71.51 & 60.70 & 40.30 & 2.10 & 96.60\% \\
\flashvid{} \venueyearlink{ICLR}{2026}{2026flashvid} & \underline{57.73} & \underline{79.64} & \underline{57.26} & \underline{73.07} & \underline{61.97} & \underline{41.19} & \textbf{2.14} & \underline{98.56\%} \\
\learnpruner{} \venueyearlink{ICLR}{2026}{takezoe2026learnpruner} & 54.67 & 75.02 & 54.09 & 68.83 & 58.72 & 38.86 & 2.02 & 93.08\% \\
\cellcolor{oursburgundy}\textbf{FoveaTree (Ours)} & \cellcolor{oursburgundy}\textbf{58.26} & \cellcolor{oursburgundy}\textbf{80.58} & \cellcolor{oursburgundy}\textbf{58.27} & \cellcolor{oursburgundy}\textbf{74.12} & \cellcolor{oursburgundy}\textbf{62.88} & \cellcolor{oursburgundy}\textbf{41.86} & \cellcolor{oursburgundy}\underline{2.12} & \cellcolor{oursburgundy}\textbf{99.61\%} \\
\hline
\end{tabular}%
}}
\end{minipage}
\end{table*}

\subsection{Additional Analysis}
\label{app:ablations}

\paragraph{Efficiency on additional backbones.}
Table~\ref{tab:app-efficiency-backbones} extends the efficiency analysis of the main paper to Qwen3-VL, Qwen2.5-VL, and LLaVA-OneVision. For each backbone, it reports total inference time, peak memory, latency, and TFLOPs together with Video-MME accuracy at 30\% and 10\% retention, measured under the identical protocol: the same Video-MME inputs, hardware, and timing procedure as the main efficiency analysis.

\begin{table}[H]
\centering
\begin{minipage}[t]{\textwidth}
\centering
\caption{Efficiency analysis at 30\% and 10\% visual-token retention on the additional backbones on Video-MME.}
\label{tab:app-efficiency-backbones}
{\scriptsize
\renewcommand{\arraystretch}{1.05}
\renewcommand{\multirowsetup}{\centering}
\setlength{\extrarowheight}{0.5pt}
\setlength{\tabcolsep}{0pt}
\resizebox{\linewidth}{!}{%
\begin{tabular}{>{\raggedright\arraybackslash}m{1.22in}|*{5}{>{\centering\arraybackslash}m{0.42695in}}|*{5}{>{\centering\arraybackslash}m{0.42695in}}}
\hline
\textbf{Method} & \textbf{Time} & \textbf{Mem } & \textbf{Latency} & \textbf{TFLOPs} & \textbf{Acc.} & \textbf{Time} & \textbf{Mem} & \textbf{Latency} & \textbf{TFLOPs} & \textbf{Acc.} \\
\hline
\rowcolor{gray!15}
\mbox{Qwen3-VL (5,753 Tokens)} & 41:24 & 18.26 & 0.920 & 77.247 & 66.15 & 41:24 & 18.26 & 0.920 & 77.247 & 66.15 \\
\hline
& \multicolumn{5}{c|}{\color{black}\textit{Retain 1,725 Tokens ($\downarrow$ \textbf{70.0\%})}} & \multicolumn{5}{c}{\color{black}\textit{Retain 575 Tokens ($\downarrow$ \textbf{90.0\%})}} \\
\hline
\tome{} \venueyear{ICLR}{2023} & 29:40 & \textbf{17.35} & 0.659 & \textbf{26.333} & 61.70 & 32:05 & \textbf{17.05} & 0.713 & \textbf{12.697} & 56.00 \\
\sttm{} \venueyear{ICCV}{2025} & 31:54 & 18.02 & 0.709 & 51.789 & 64.54 & 31:34 & 17.97 & 0.702 & 44.971 & 63.38 \\
\flashvid{} \venueyear{ICLR}{2026} & \underline{26:57} & \underline{17.51} & \underline{0.599} & \underline{28.835} & \underline{64.77} & \textbf{20:59} & \underline{17.08} & \textbf{0.466} & \underline{13.469} & \underline{63.95} \\
\rowcolor{oursburgundy}
\textbf{FoveaTree (Ours)} & \textbf{24:25} & 17.80 & \textbf{0.543} & 34.303 & \textbf{65.38} & \underline{22:40} & 17.62 & \underline{0.504} & 27.835 & \textbf{64.92} \\
\hline
\noalign{\vspace{6pt}}
\hline
\rowcolor{gray!15}
\mbox{Qwen2.5-VL (5,753 Tokens)} & 27:39 & 16.83 & 0.615 & 83.566 & 61.85 & 27:39 & 16.83 & 0.615 & 83.566 & 61.85 \\
\hline
& \multicolumn{5}{c|}{\color{black}\textit{Retain 1,725 Tokens ($\downarrow$ \textbf{70.0\%})}} & \multicolumn{5}{c}{\color{black}\textit{Retain 575 Tokens ($\downarrow$ \textbf{90.0\%})}} \\
\hline
\tome{} \venueyear{ICLR}{2023} & \underline{22:14} & \textbf{16.12} & \underline{0.494} & \textbf{24.708} & 56.33 & 23:52 & \textbf{15.98} & 0.530 & \textbf{9.089} & 51.43 \\
\sttm{} \venueyear{ICCV}{2025} & 25:41 & 16.72 & 0.571 & 62.544 & 60.81 & 24:05 & 16.72 & 0.535 & 56.966 & 60.04 \\
\flashvid{} \venueyear{ICLR}{2026} & 23:31 & \underline{16.17} & 0.522 & \underline{28.776} & \underline{60.99} & \textbf{16:18} & \textbf{15.98} & \textbf{0.362} & \underline{10.519} & \underline{60.61} \\
\rowcolor{oursburgundy}
\textbf{FoveaTree (Ours)} & \textbf{20:23} & 16.65 & \textbf{0.453} & 41.437 & \textbf{61.44} & \underline{19:14} & 16.65 & \underline{0.427} & 35.897  & \textbf{61.37} \\
\hline
\noalign{\vspace{6pt}}
\hline
\rowcolor{gray!15}
\mbox{LLaVA-OV (12,544 Tokens)} & 53:33 & 17.88 & 1.190 & 197.331 & 58.56 & 53:33 & 17.88 & 1.190 & 197.331 & 58.56 \\
\hline
& \multicolumn{5}{c|}{\color{black}\textit{Retain 3,776 Tokens ($\downarrow$ \textbf{70.0\%})}} & \multicolumn{5}{c}{\color{black}\textit{Retain 1,280 Tokens ($\downarrow$ \textbf{90.0\%})}} \\
\hline
\tome{} \venueyear{ICLR}{2023} & 37:01 & \textbf{16.33} & 0.823 & \textbf{53.796} & 55.13 & 37:57 & \textbf{15.87} & 0.843 & \textbf{18.576} & 49.43 \\
\sttm{} \venueyear{ICCV}{2025} & 46:59 & 17.64 & 1.044 & 146.065 & 57.49 & 46:44 & 17.64 & 1.039 & 133.487 & 56.54 \\
\flashvid{} \venueyear{ICLR}{2026} & \underline{33:51} & \underline{16.47} & \underline{0.752} & \underline{63.186} & \underline{58.37} & \textbf{18:51} & \underline{16.08} & \textbf{0.419} & \underline{21.631} & \underline{57.73} \\
\rowcolor{oursburgundy}
\textbf{FoveaTree (Ours)} & \textbf{31:01} & 17.77 & \textbf{0.689} & 91.245 & \textbf{58.46} & \underline{28:22} & 17.77 & \underline{0.630} & 80.568  & \textbf{58.26} \\
\hline
\end{tabular}%
}}
\end{minipage}
\end{table}

\begin{wraptable}{l}{3.1465in}
\centering
\caption{Ablation on the maximum quadtree depth $D$ at 30\% retention. Latency follows the efficiency protocol of the main paper.}
\label{tab:app-quadtree-depth}
{\fontsize{6.856pt}{8.771pt}\selectfont
\renewcommand{\arraystretch}{1.0}%
\setlength{\extrarowheight}{0.452pt}%
\setlength{\tabcolsep}{0pt}%
\resizebox{3.1465in}{!}{%
\begin{tabular}{>{\centering\arraybackslash}m{0.82in}|>{\centering\arraybackslash}m{0.50in}>{\centering\arraybackslash}m{0.50in}>{\centering\arraybackslash}m{0.50in}>{\centering\arraybackslash}m{0.50in}}
\hline
\textbf{Depth $D$} & \textbf{VideoMME} & \textbf{VDC$_{\mathrm{Acc.}}$} & \textbf{Avg.} & \textbf{Lat.} \\
\hline
$D=2$ & 63.87 & 38.92 & 96.71\% & \textbf{1.871} \\
$D=3$ & 65.24 & 40.31 & 99.47\% & \underline{1.942} \\
\rowcolor{oursburgundy}\cellcolor{oursburgundy}\textbf{$D=4$} & \cellcolor{oursburgundy}\textbf{66.01} & \cellcolor{oursburgundy}\textbf{41.39} & \cellcolor{oursburgundy}\textbf{101.40\%} & \cellcolor{oursburgundy}2.006 \\
$D=5$ & \underline{65.83} & \underline{41.02} & \underline{100.80\%} & 2.089 \\
\hline
\end{tabular}%
}}
\end{wraptable}

\paragraph{Quadtree depth.}
These supplementary ablations are conducted on the InternVL3.5-8B backbone, i.e., the primary model of the main paper. The depth setting controls the MGDE hierarchy and the spatial extent of its root regions. This ablation varies the maximum depth $D\in\{2,3,4,5\}$, where $D=4$ corresponds to the default $8\rightarrow4\rightarrow2\rightarrow1$ schedule, while $D=5$ starts from a single $16\times16$ root covering the entire frame grid and follows the $16\rightarrow8\rightarrow4\rightarrow2\rightarrow1$ schedule. Table~\ref{tab:app-quadtree-depth} reports Video-MME accuracy, VDC accuracy, and the average relative performance, together with latency measured under the efficiency protocol of the main paper.

\begin{wraptable}{l}{2.6675in}
\centering
\caption{Ablation on the ATP split-score composition at 30\% retention.}
\label{tab:app-score-composition}
{\fontsize{6.856pt}{8.771pt}\selectfont
\renewcommand{\arraystretch}{1.0}%
\setlength{\extrarowheight}{0.452pt}%
\setlength{\tabcolsep}{0pt}%
\resizebox{2.6675in}{!}{%
\begin{tabular}{>{\centering\arraybackslash}m{0.9in}|>{\centering\arraybackslash}m{0.50in}>{\centering\arraybackslash}m{0.50in}>{\centering\arraybackslash}m{0.50in}}
\hline
\textbf{$\kappa_u$} & \textbf{VideoMME} & \textbf{VDC$_{\mathrm{Acc.}}$} & \textbf{Avg.} \\
\hline
$\kappa_u=\delta_u$ & \underline{65.18} & \underline{40.42} & \underline{99.57\%} \\
$\kappa_u=\rho_u$ & 64.15 & 39.47 & 97.61\% \\
$\kappa_u=a_u$ & 63.36 & 38.71 & 96.07\% \\
\rowcolor{oursburgundy}\cellcolor{oursburgundy}$\kappa_u=\delta_u+\rho_u$ & \cellcolor{oursburgundy}\textbf{66.01} & \cellcolor{oursburgundy}\textbf{41.39} & \cellcolor{oursburgundy}\textbf{101.40\%} \\
\hline
\end{tabular}%
}}
\end{wraptable}

\paragraph{Score composition.}
The ATP split priority combines feature heterogeneity with propagated attention relevance, $\kappa_u=\delta_u+\rho_u$. This ablation isolates the two terms: ranking nodes by heterogeneity alone ($\kappa_u=\delta_u$) preserves content-diverse regions but ignores the query, whereas ranking by relevance alone ($\kappa_u=\rho_u$) follows attention without accounting for the internal diversity of a region. We further include the uncalibrated variant $\kappa_u=a_u$, where $a_u=\max_{v\in\V(u)}a_v$ propagates the raw averaged leaf attention defined in the main paper. This bypasses the within-unit percentile rescaling and isolates the contribution of that calibration step. Table~\ref{tab:app-score-composition} compares the four configurations under the same retention budget. The combined score retains the strongest results on both metrics, indicating that the two terms are complementary: relevance locates query-related branches while heterogeneity prevents merging internally diverse regions.
\begin{figure}[H]
\centering
\includegraphics[width=\linewidth]{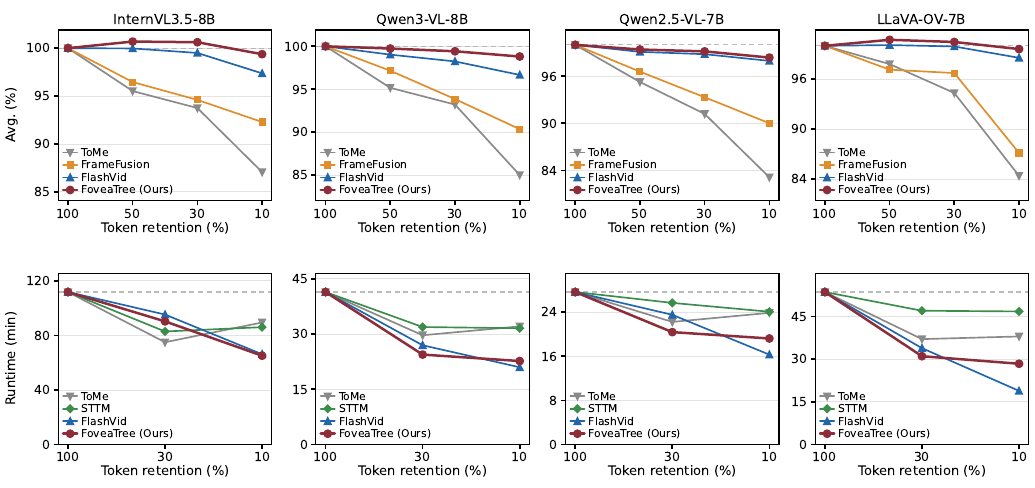}
\caption{Cross-backbone comparison. Top: average relative performance versus the visual-token retention ratio on InternVL3.5-8B, Qwen3-VL-8B, Qwen2.5-VL-7B, and LLaVA-OneVision-7B for FoveaTree, FlashVID, FrameFusion, and ToMe; the dashed line marks the unpruned reference. Bottom: total Video-MME runtime under the efficiency protocol of the main paper for FoveaTree, FlashVID, STTM, and ToMe; the dashed line marks the runtime of the unpruned model.}
\label{fig:app-backbone-curves}
\end{figure}

\paragraph{Cross-backbone result analysis.}
Figure~\ref{fig:app-backbone-curves} summarizes performance retention and runtime across four backbones. FoveaTree maintains strong performance across these architectures while reducing inference time relative to the unpruned models. At 30\% retention, it achieves both the highest Video-MME accuracy and the lowest latency among the methods in Table~\ref{tab:app-efficiency-backbones} on Qwen3-VL, Qwen2.5-VL, and LLaVA-OneVision. On InternVL3.5, it is more accurate and faster than FlashVID and LearnPruner. These results demonstrate a favorable accuracy--latency trade-off across backbones. At 10\% retention, it preserves the highest Video-MME accuracy among these methods, while FlashVID is faster on the Qwen and LLaVA backbones.

The results also suggest that backbones with larger original visual-token sequences can offer more room for compression. This is particularly evident on InternVL3.5, whose 16,384-token input can be compressed at both 50\% and 30\% retention while exceeding the unpruned reference in average relative performance. These gains show that retaining more visual tokens does not necessarily improve understanding. We hypothesize that excessive visual redundancy, together with a model's limited ability to use long visual contexts effectively, can reduce the benefit of additional tokens; structured compression may alleviate this burden by retaining informative regions. The cross-backbone observations are consistent with this explanation, although differences in architecture and preprocessing prevent attributing the effect to token count alone.

\section{Evaluation Benchmarks}
\label{app:settings}
We use the official test splits, subject to the subset specifications below.

\paragraph{Video-MME.}
Video-MME~\citep{fu2024videomme} evaluates video understanding across short, medium, and long videos. Its questions cover perception and reasoning in diverse content domains. We evaluate without subtitles, report answer accuracy, and use the dataset for efficiency measurements with a 64-frame cap. The fine-grained analysis examines OCR, counting, and object-related questions, which place different demands on the retained visual content.

\paragraph{NExT-QA.}
NExT-QA~\citep{xiao2021nextqa} includes descriptive, temporal, and causal questions about video events. In addition to recognizing the visible content, the model must connect actions and explain their temporal or causal relationships. We report question-answering accuracy to assess whether token pruning preserves the evidence required for these judgments.

\paragraph{MVBench.}
MVBench~\citep{li2024mvbench} organizes video understanding into multiple tasks involving actions, objects, spatial relationships, and temporal changes. Its diversity complements the broad evaluation in Video-MME. We use the benchmark's accuracy metric and retain the original evaluation protocol when comparing the full model and compressed models. The official release contains 4{,}000 samples, but 200 samples of the Fine-grained Pose task are missing from the version we evaluate, so all reported numbers use the remaining 3{,}800 samples. Every method, including the uncompressed reference, is evaluated on the identical 3{,}800-sample subset, ensuring a fair comparison.

\paragraph{Valor32k.}
The VALOR dataset~\citep{chen2023valor} supports vision, audio, and language understanding. We evaluate on the visual subset of Valor32k using the visual question-answering setting. This setting measures the model's ability to answer questions using video content and complements the action-oriented evaluation in NExT-QA and MVBench.

\paragraph{VNBench.}
VNBench~\citep{zhao2025vnbench} evaluates retrieval, ordering, and counting through the Video Needle In A Haystack construction. Relevant visual events are embedded within longer sequences, making the benchmark useful for assessing temporal localization and the retention of sparse evidence. We report accuracy under the same input configuration for the full and compressed models.

\paragraph{VDC.}
VDC, introduced with AuroraCap~\citep{chai2025auroracap}, evaluates detailed video captioning using structured reference descriptions. We follow the official VDC QA evaluation pipeline, using Llama-3.1-8B-Instruct for semantic yes/no judgments. We report Acc. and Score, denoted $VDC_{\mathrm{Acc.}}$ and $VDC_{\mathrm{Score}}$, respectively. Detailed captioning requires the model to describe several aspects of the scene within one response and provides a complementary evaluation to video question answering. Subsequent work such as VDC-Agent~\citep{wang2025vdcagent} also uses this benchmark to study detailed caption generation.

\paragraph{Average relative performance.}
Let $s_j$ and $s_j^{\mathrm{Full}}>0$ denote a method's score and the full model's score on reported metric $j$. For a table with $J$ task metrics, Avg. is the equal-weighted mean $\mathrm{Avg.}=100J^{-1}\sum_{j=1}^{J}s_j/s_j^{\mathrm{Full}}$. Every reported task metric contributes once. Thus, $VDC_{\mathrm{Acc.}}$ and $VDC_{\mathrm{Score}}$ contribute separately when both appear, and the set of metrics changes when a table reports a smaller benchmark subset. Memory, computation, and latency are excluded from this mean. This normalization makes metrics with different numerical scales comparable without assigning larger weights to benchmarks with more samples.

For context, MMBench and MM-Vet provide complementary multimodal evaluation protocols~\citep{liu2024mmbench,yu2024mmvet}. We do not substitute these datasets for the official benchmark splits used here.

\section{Implementation Details}
\label{app:efficiency}

\subsection{Reproduction Details of Compared Baselines}
\label{app:baseline-reproduction}
We follow the configurations and token-retention settings provided in the public code repositories of the compared methods. All baselines are evaluated using the same benchmark pipeline as FoveaTree, with identical input preprocessing, frame sampling, decoding settings, and hardware.

\subsection{Reproduction Details of FoveaTree}
\label{app:foveatree-reproduction}
Throughout the main paper, per-frame visual grids, attention normalization, and token budgets refer to the backbone's native temporal units: one sampled frame for InternVL3.5 and LLaVA-OneVision, but two sampled frames per unit for Qwen3-VL and Qwen2.5-VL.
FoveaTree uses the public checkpoints of InternVL3.5-8B, Qwen3-VL-8B, Qwen2.5-VL-7B, and LLaVA-OneVision-7B without any additional training, calibration, or reference passes. MGDE operates on layer-0 visual features with a root side length of 8 and a merge-gain threshold of 0.005; the per-unit leaf caps follow Eq.~\eqref{eq:app-mgde-capacity} with $L_{\max}=196$ for InternVL3.5 and LLaVA-OneVision, $L_{\max}=H_tW_t$ for Qwen2.5-VL, and $L_{\max}=120$ for Qwen3-VL. ATP is inserted at decoder layer 18 with the default split score $\kappa_u=\delta_u+\rho_u$ and a maximum quadtree depth of $D=4$ (Appendix~\ref{app:ablations}); the same configuration is transferred unchanged to Qwen2.5-VL and LLaVA-OneVision. Inference uses the same batch-one bfloat16, SDPA, and greedy-decoding configuration as the baselines, with \texttt{max\_new\_tokens}=512 for VDC and 8 for all other benchmarks. Efficiency measurements synchronize CUDA timing and exclude warm-up iterations, and all token counts are audited per temporal unit as described in Appendix~\ref{app:budget-control}.

\subsection{Token Budget Alignment}
\label{app:budget-control}

\paragraph{Budget definition.}
Let $N_t$ be the number of original visual tokens of a native temporal unit $t$, and let $r\in\{50\%,30\%,10\%\}$ be the nominal retention ratio. The per-unit budget and the final output count are defined as
\begin{equation}
B_t=\min\!\bigl(N_t,\lceil rN_t\rceil\bigr),
\qquad
K_t=\min(B_t,M_t),
\label{eq:app-budget}
\end{equation}
where $M_t$ is the number of tokens surviving MGDE and $K_t$ is the number of tokens finally emitted by ATP. By construction $K_t\leq B_t$: the budget is a hard upper bound, not a quota that must be filled. Text tokens are excluded from $N_t$, and the budget is computed from the original visual grid of each temporal unit, so backbones with dynamic resolution obtain per-unit budgets that follow their actual token counts. Qwen3-VL and Qwen2.5-VL use \texttt{temporal\_patch\_size=2}, so 64 sampled frames form 32 native temporal units per question-video input; the spatial token count of a Qwen2.5-VL unit follows its dynamic resolution. InternVL3.5 and LLaVA-OneVision retain one temporal unit per sampled frame.

\paragraph{Unit-local budget resolution.}
When $M_t>B_t$, ATP applies a unit-local budget resolver. (i) The frontier is initialized with the unit-level routing node, to which the surviving roots of the restricted MGDE forests are attached. (ii) At each step, the expandable node with the highest split priority $\kappa_u$ is selected. (iii) If a node has $k$ valid children, a complete expansion increases the frontier size by $k-1$, and the expansion is applied whenever this increase fits within the remaining budget. (iv) If a complete expansion would cross the budget, let $F$ be the current frontier size and $d=B_t-F$ the remaining slots. ATP replaces the selected node with its $d$ highest-priority children and one \emph{residual cluster} containing all remaining children, yielding $F-1+d+1=B_t$ frontier entries. (v) Every terminal cluster maps to exactly one output token, so the resolver terminates with exactly $B_t$ outputs. Because quadtree children are spatially disjoint, the terminal clusters form a non-overlapping frontier with unique spatial supports, and restoring temporal-unit and raster order preserves the original video sequence. The residual-cluster mechanism absorbs the granularity of quadtree expansion: exposing a node adds several tokens at once and could otherwise overshoot the budget, whereas a residual cluster never merges subtrees from different parents.

\paragraph{Under-budget behavior.}
When MGDE leaves fewer tokens than the requested ATP budget, FoveaTree keeps all MGDE survivors and does not synthesize, pad, or borrow tokens: no tokens are borrowed from other temporal units, the MGDE selection itself is not altered, and the final output is $K_t=M_t$. Algorithm~\ref{alg:app-budget} summarizes the complete per-unit procedure, and the implementation enforces the contract with a hard check, $\mathrm{count}=\min(B_t,M_t)$, on every temporal unit; the InternVL3.5, Qwen3-VL, Qwen2.5-VL, and LLaVA-OneVision pipelines raise an immediate error instead of silently accepting a violation.

\begin{algorithm}[H]
\caption{Unit-local token budget resolution for one temporal unit.}
\label{alg:app-budget}
\begin{algorithmic}[1]
\State $N_t\gets$ number of original visual tokens;\quad $B_t\gets\min(N_t,\lceil rN_t\rceil)$
\State $M_t\gets$ number of tokens after MGDE
\If{$M_t\leq B_t$}
\State emit all MGDE survivors and set $K_t\gets M_t$
\Else
\State frontier $\gets$ \{unit-level routing node\}
\While{$\abs{\mathrm{frontier}}<B_t$}
\State $u\gets$ expandable node with the highest split priority $\kappa_u$
\If{a complete expansion of $u$ fits within $B_t$}
\State replace $u$ by all of its valid children
\Else
\State $d\gets B_t-\abs{\mathrm{frontier}}$
\State replace $u$ by its $d$ highest-priority children and one residual cluster
\State the residual cluster contains all remaining children of $u$
\State \textbf{break}
\EndIf
\EndWhile
\State emit the frontier clusters and set $K_t\gets\abs{\mathrm{frontier}}$
\EndIf
\State \textbf{assert} $K_t=\min(B_t,M_t)$
\end{algorithmic}
\end{algorithm}

\begin{wraptable}{R}{2in}
\vspace{-8pt}
\centering
\caption{Definition of the per-unit token-budget symbols.}
\label{tab:app-budget-definitions}
{\fontsize{6.856pt}{8.771pt}\selectfont
\renewcommand{\arraystretch}{1.0}%
\setlength{\extrarowheight}{0.452pt}%
\setlength{\tabcolsep}{0pt}%
\resizebox{2in}{!}{%
\begin{tabular}{>{\centering\arraybackslash}m{0.4in}|>{\raggedright\arraybackslash}m{1.5in}}
\hline
\textbf{Symbol} & \textbf{Meaning} \\
\hline
$N_t$ & number of original visual tokens \\
$r$ & nominal retention ratio \\
$B_t$ & ATP budget upper bound, $\min(N_t,\lceil rN_t\rceil)$ \\
$M_t$ & number of tokens surviving MGDE \\
$K_t$ & number of tokens finally emitted by ATP \\
$B_t-K_t$ & budget left unfilled by MGDE \\
$K_t-B_t$ & signed budget excess, must be $\leq 0$ \\
\hline
\end{tabular}%
}}
\vspace{-8pt}
\end{wraptable}

\paragraph{MGDE tree capacity.}
MGDE applies a leaf-count cap to the entire quadtree forest of each temporal unit, rather than separately to each root. Let $L_t^{\mathrm{tree}}$ denote its current number of terminal regions. A candidate split is permitted only if the resulting forest satisfies
\begin{equation}
L_t^{\mathrm{tree}}\leq L_{\max}.
\label{eq:app-mgde-capacity}
\end{equation}
We use $L_{\max}=196$ for InternVL3.5 and LLaVA-OneVision, $L_{\max}=H_tW_t$ for Qwen2.5-VL, and $L_{\max}=120$ for Qwen3-VL, where $H_t\times W_t$ is the unit's visual grid. The cap is an upper bound, not a prescribed leaf count: the split threshold and forest structure determine where refinement stops. It is distinct from the ATP output budget $B_t$. Subsequent density-guided deletion and merging can further reduce the number of survivors $M_t$, producing the unfilled budgets reported below.

\paragraph{Budget auditing.}
Table~\ref{tab:app-budget-definitions} summarizes the notation. Table~\ref{tab:app-budget-audit} audits the actual token counts over the full Video-MME evaluation for each backbone: InternVL3.5, Qwen3-VL, Qwen2.5-VL, and LLaVA-OneVision. Each evaluation covers 2{,}700 questions over 900 videos. For InternVL3.5, the audit contains 172{,}800 temporal units across 2{,}700 question-video inputs. On this run, MGDE leaves a mean of 187.38 survivors per 256-token temporal unit, the share of units below budget is 4.45\%, 1.93\%, and 0.52\% at 50\%, 30\%, and 10\% retention, and all 172{,}800 InternVL3.5 units satisfy $K_t=\min(B_t,M_t)$ with zero over-budget units. Across all audited backbones, no temporal unit ever exceeds its budget; every deviation is an unfilled budget, i.e., a case in which MGDE produces fewer survivors than the budget allows. Each input's total budget is the sum of its per-unit budgets; for instance, 30\% retention on LLaVA-OneVision yields $64\times\lceil 0.30\times196\rceil=3{,}776$ tokens. Unit-level and input-level statistics are kept separate: the large input-level deficits, such as 3{,}144 tokens at 50\% retention, aggregate the per-unit under-fills across the temporal units in an input and are not deficits of any single unit.

\begin{table}[H]
\centering
\caption{Token-budget audit over all 2,700 Video-MME question-video inputs for each backbone. Actual is the mean final visual-token count per input; Range is its minimum and maximum. Unfilled and Max unfilled are the mean and maximum of each input's own budget minus its final token count. Util. is total retained tokens divided by total budget. Units counts native temporal units across all inputs; Over-budget counts units exceeding their own budgets. Budgets are computed separately for each input, including dynamic-resolution inputs. }
\label{tab:app-budget-audit}
\label{tab:app-budget-overshoot}
\setlength{\tabcolsep}{3pt}
\renewcommand{\arraystretch}{1.15}
\resizebox{\linewidth}{!}{%
\begin{tabular}{lrrrrrrrr}
\toprule
Model & Ret. & Units & Actual & Range & Unfilled & Max unfilled & Util. & Over-budget \\
\midrule
InternVL3.5 & 50\% & 172,800 & 8,043.21 & 5,048--8,192 & 148.79 & 3,144 & 98.2\% & 0 \\
InternVL3.5 & 30\% & 172,800 & 4,881.93 & 3,591--4,928 & 46.07 & 1,337 & 99.1\% & 0 \\
InternVL3.5 & 10\% & 172,800 & 1,656.70 & 1,326--1,664 & 7.30 & 338 & 99.6\% & 0 \\
\midrule
LLaVA-OneVision & 30\% & 172,800 & 3,628.93 & 1,685--3,776 & 147.07 & 2,091 & 96.105\% & 0 \\
LLaVA-OneVision & 10\% & 172,800 & 1,251.26 & 795--1,280 & 28.74 & 485 & 97.755\% & 0 \\
\midrule
Qwen3-VL & 30\% & 86,400 & 1,393.19 & 956--1,440 & 12.85 & 452 & 99.086\% & 0 \\
Qwen3-VL & 10\% & 86,400 & 477.97 & 375--480 & 1.31 & 105 & 99.726\% & 0 \\
\midrule
Qwen2.5-VL & 30\% & 86,400 & 1,716.51 & 1,318--1,888 & 9.46 & 410 & 99.452\% & 0 \\
Qwen2.5-VL & 10\% & 86,400 & 574.09 & 450--640 & 1.34 & 126 & 99.767\% & 0 \\
\bottomrule
\end{tabular}%
}
\end{table}

\paragraph{Fairness of the comparison.}
All methods are evaluated under the same nominal maximum output-token budget. \tome{}, \sttm{}, and \flashvid{} fill this budget exactly, whereas FoveaTree enforces it as a hard upper bound and retains all MGDE survivors without padding when fewer are available. This is a same-maximum-budget comparison, not a per-sample identical-token-count comparison; Table~\ref{tab:app-budget-audit} reports FoveaTree's actual budget utilization. An equal-or-lower final token count does not by itself imply an equal-or-lower total inference cost, because methods differ in their pruning locations and overhead. FLOPs are computed for the actual execution, accounting for the sequence lengths processed at each stage and the pruning operations; latency and peak memory are measured over the full inference pipeline, including tree construction, attention extraction, and pruning.

\subsection{Properties of Tree-Based Aggregation}
\label{app:aggregation-properties}

\paragraph{Weighted merging of terminal clusters.}
For a terminal cluster $C$, the weights $w_v=(\rho_v+\epsilon)/\sum_{v'\in C}(\rho_{v'}+\epsilon)$ are positive and sum to one. The merged hidden state $\widehat h_C^\ell=\sum_{v\in C}w_vh_v^\ell$ lies in the convex hull of its constituent states. The constituent with the highest attention provides the position anchor, and the merged state retains that constituent's original position under RoPE~\citep{su2024roformer}. In particular, $\lVert\widehat h_C^\ell\rVert_2\leq\max_{v\in C}\lVert h_v^\ell\rVert_2$. For a constituent state $h_i^\ell$, the triangle inequality also gives $\lVert h_i^\ell-\widehat h_C^\ell\rVert_2\leq\sum_{v\in C}w_v\lVert h_i^\ell-h_v^\ell\rVert_2$. These relations explain why homogeneous groups admit a compact representation and motivate combining heterogeneity with attention when deciding to merge.

\paragraph{Stability of bottom-up propagation.}
For a fixed surviving tree, consider two sets of normalized leaf scores $\rho$ and $\rho'$. At any internal node $u$, maximum propagation satisfies
\begin{equation}
\left|\max_{v\in\V(u)}\rho_v-\max_{v\in\V(u)}\rho'_v\right|
\leq\max_{v\in\V(u)}|\rho_v-\rho'_v|.
\label{eq:app-max-stability}
\end{equation}
To see this, let $v^*$ maximize the first score set. The first maximum minus the second is at most $\rho_{v^*}-\rho'_{v^*}$. Exchanging the two score sets gives the reverse bound. Propagating a maximum therefore does not amplify the largest perturbation of the normalized leaf scores on a fixed tree.

Moreover, $\rho_u\geq\rho_v$ for every surviving descendant $v$. A region containing strong evidence cannot receive a smaller propagated relevance merely because it also contains weakly attended units. This property explains the preservation of regional support under maximum aggregation. It applies after the within-unit percentile normalization and does not establish a corresponding bound on the raw attention distribution or on the final answer accuracy.

\section{Related Work}
\label{app:related}

\paragraph{Visual representations and video LLMs.}
Vision Transformers~\citep{dosovitskiy2021vit} and hierarchical variants such as Swin~\citep{liu2021swin} encode images as patch sequences; CLIP and SigLIP~\citep{radford2021clip,zhai2023siglip} use language supervision; MAE, DINOv2, and DINO~\citep{he2022mae,oquab2023dinov2,caron2021dino} use self-supervision; and BLIP, BLIP-2, Flamingo, and LLaVA~\citep{li2022blip,li2023blip2,alayrac2022flamingo,liu2023llava,liu2024llava15} connect visual encoders to language models. Video-LLaVA, Video-ChatGPT, Chat-UniVi, TimeChat, LLaMA-VID, and InternVideo2~\citep{lin2024videollava,maaz2024videochatgpt,jin2024chatunivi,ren2024timechat,li2024llamavid,wang2024internvideo2} extend multimodal modeling to video; Qwen-VL, Qwen2-VL, and LLaVA-UHD~\citep{bai2023qwenvl,wang2024qwen2vl,xu2024llavauhd} accommodate varying image resolutions and aspect ratios; and earlier InternVL releases and ShareGPT4V~\citep{chen2024internvl,chen2024internvl15,chen2024sharegpt4v} add open-source resources for visual instruction tuning. We evaluate InternVL3.5-8B, Qwen3-VL-8B, Qwen2.5-VL-7B, and LLaVA-OneVision-7B~\citep{wang2025internvl35,bai2025qwen3vl,bai2025qwen25vl,li2025llavaonevision}.

\paragraph{Token selection and compression.}
TokenLearner learns compact visual tokens~\citep{ryoo2021tokenlearner}, DynamicViT and ATS adaptively select tokens~\citep{rao2021dynamicvit,fayyaz2022ats}, and EViT and Token Pooling consolidate redundant information~\citep{liang2022evit,marin2023tokenpooling}. In VLMs, LLaVA-PruMerge combines selection and merging~\citep{shang2024prumerge}, HiRED allocates image-partition budgets~\citep{arif2024hired}, and VisPruner balances saliency and diversity~\citep{zhang2024fastervlm}; ZipVL performs dynamic token sparsification~\citep{he2024zipvl}, and FocusLLaVA and DeCo study coarse-to-fine or decoupled compression~\citep{zhu2024focusllava,yao2024deco}. PyramidDrop, SparseVLM, and ATP-LLaVA prune across decoder layers~\citep{xing2024pyramiddrop,zhang2024sparsevlm,ye2025atpllava}; MADTP uses multimodal alignment~\citep{cao2024madtp}; and DART, FEATHER, HoloV, and MustDrop target duplicate features, spatial coverage, holistic context, and training-free reduction~\citep{wen2025dart,endo2024feather,zou2025holov,han2024mustdrop}. Our baselines include ToMe's similarity merging~\citep{bolya2023tome}, FastV's decoder attention pruning~\citep{chen2024fastv}, VisionZip's dominant and contextual selection~\citep{yang2025visionzip}, FrameFusion's combined similarity and importance~\citep{fu2025framefusion}, STTM's multi-granular spatiotemporal merging~\citep{hyun2025sttm}, FlashVID's attention- and diversity-based tree merging~\citep{2026flashvid}, and LearnPruner's learned module with decoder attention pruning~\citep{takezoe2026learnpruner}. Attention sinks~\citep{xiao2024streamingllm} and position-sensitive retrieval~\citep{liu2024lostmiddle} are documented in LLMs; VLM debiasing methods address biased token importance~\citep{zhao2025attentiondebiasing,zhang2025d2pruner}. FoveaTree reuses distortion-based spatial quadtrees for query-guided pruning without training.

Related analyses cover visual shortcomings and internal processing in multimodal models~\citep{tong2024eyes,neo2024interpreting}, high-resolution processing~\citep{li2024minigemini}, and the Transformer attention underlying decoder-side selection~\citep{vaswani2017attention}.

\paragraph{Captioning semantics, temporal evidence, and reliability.}
Audio-Visual Exchange-Aware Token Pruning studies efficient audio-visual captioning~\citep{meng2026avpruning}, providing a task-specific connection between token reduction and language generation. Complementary captioning work explores structured concept prediction~\citep{wang2023structuredconcepts} and Transformer-based graphs of visual relationships~\citep{li2024visualrelationships}. These directions motivate examining whether compression preserves objects and their relations, beyond reporting the number of retained tokens. Caption coherence has also been explored for video summarization~\citep{ye2025captioncoherence}, while active moment discovery addresses partially relevant video retrieval~\citep{song2025activemoment}; both concern the selection or organization of temporal evidence.

Emotional and subjective video captioning extends the semantic content of descriptions through dual-path collaborative generation~\citep{ye2024emotionalcaptioning}, multi-round emotion--cause pair extraction~\citep{ye2025emotioncause}, and subjective--objective emotion-correlated generation~\citep{chen2026subjectivecaptioning}. FACE-net studies factual calibration and emotion augmentation in retrieval-enhanced emotional video captioning~\citep{chen2026facenet}, and recent work evaluates and mitigates emotion hallucinations in MLLMs~\citep{zeng2026emotionhallucinations}. These tasks highlight additional aspects of description quality that a compression method could affect. Our current evaluations measure general video question answering and captioning; they do not establish performance on these specialized emotional-captioning tasks or reductions in emotion hallucinations.

\paragraph{Additional related studies.}
Additional work studies affective explanation captioning with consensus-prompted emotion reasoning~\citep{song2026affectiveexplanation}, fine-grained emotion--cause extraction for video captions~\citep{chen2026emotioncause}, and sentiment-oriented generation for live video comments~\citep{fu2024livecommenting}. Earlier multimodal studies address weakly supervised actor--action segmentation~\citep{chen2023actoraction}, contour-augmented concept prediction for image captioning~\citep{wang2023contourcaption}, and compressed-video referring object segmentation~\citep{chen2022compressedreferring}. These studies address caption semantics and language-guided visual grounding, complementing the token-compression literature discussed above.

\section{More Visualizations}
\label{app:visualizations}

We visualize the spatial compression performed by MGDE and ATP and compare the responses of InternVL3.5 and Qwen3-VL with and without FoveaTree. In the compressed-frame displays, filled grid cells mark retained tokens, small inset patches represent merged regions, and empty regions indicate removed content.

\subsection{Visualizations of MGDE}
\label{app:properties}

Each figure arranges input frames, MGDE density maps, and compressed representations from top to bottom. Yellow, orange, and magenta mark coarse, medium, and fine terminal regions, i.e., dropped, merged, and retained tokens. MGDE assigns finer partitions to detailed regions and coarser ones to homogeneous regions.

\begin{figure}[H]
\centering
\includegraphics[width=0.74\linewidth]{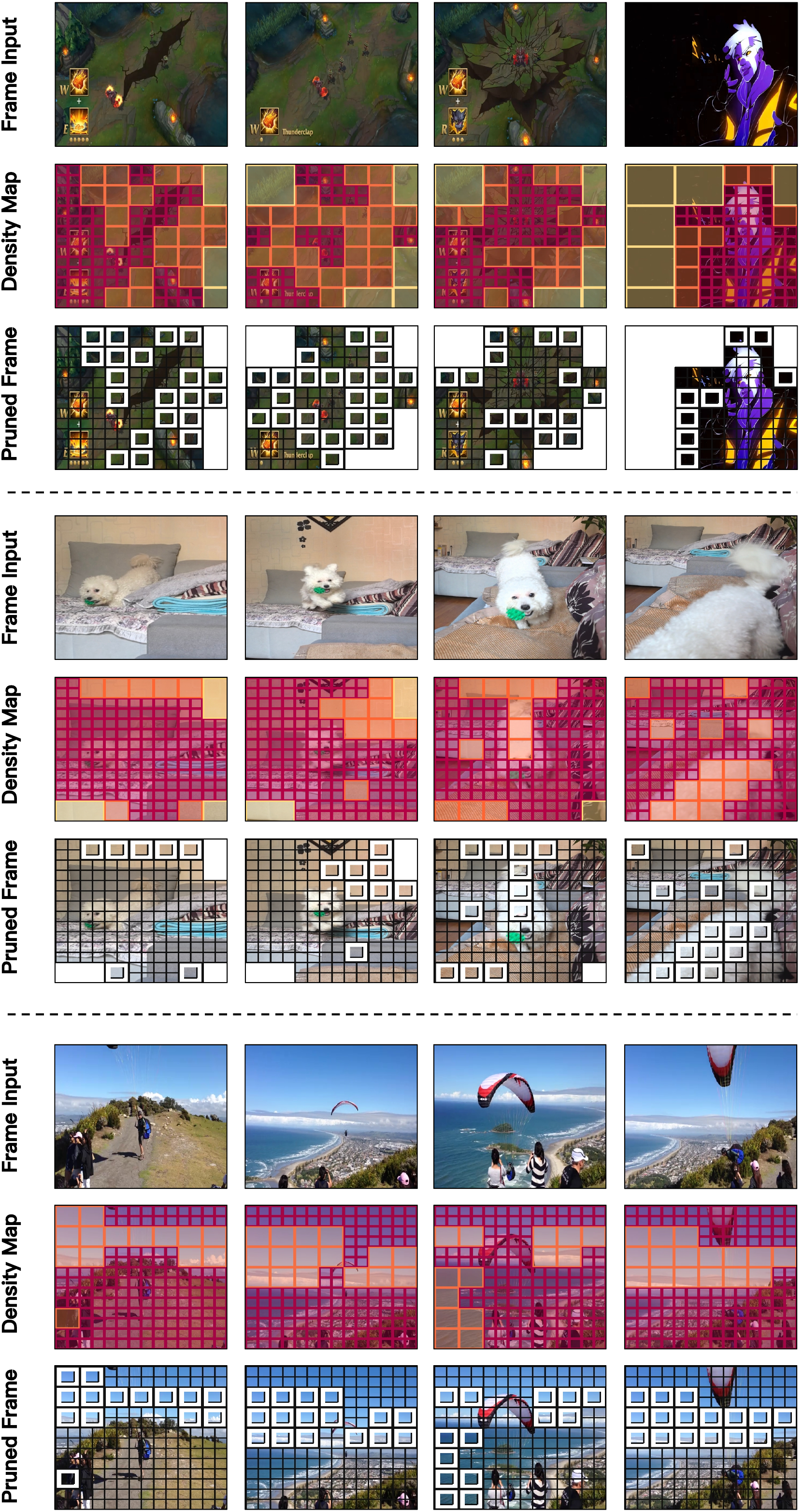}

\end{figure}

\subsection{Visualizations of ATP}
\label{app:atp-viz}

Each example shows the query above four sampled frames, with the MGDE output in the upper row and the final ATP output in the lower row. The queries concern a hand gesture, the number of athletes in a tennis match, and a summary of a manual activity. ATP uses the query and inherited spatial hierarchy to further compress the MGDE survivors. The final representations retain regions around the girl's face and hands, people in the tennis scene, and the hands and objects involved in the demonstrated activity, respectively, illustrating the spatial allocation of the final token budget.

\begin{figure}[H]
\centering
\includegraphics[width=0.9\linewidth]{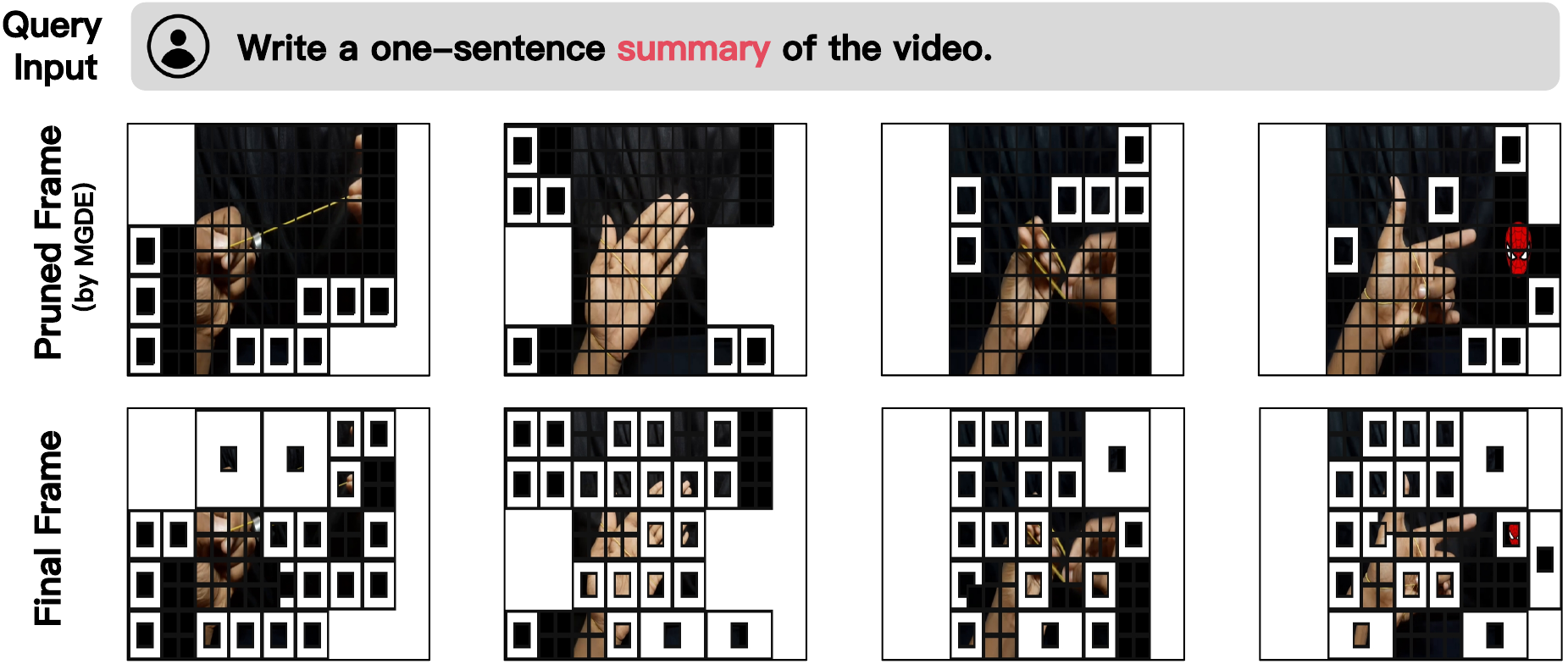}\par
\vspace{5pt}
\includegraphics[width=0.9\linewidth]{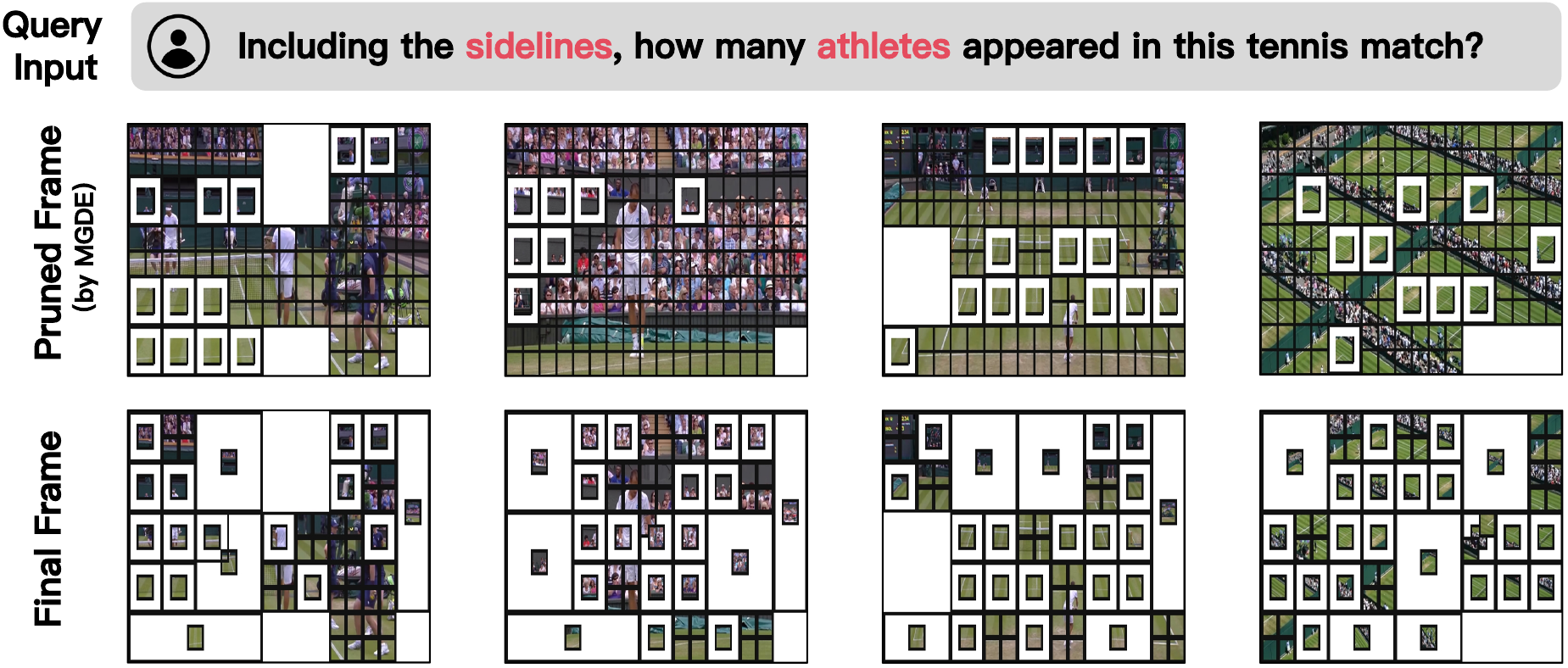}\par
\includegraphics[width=0.9\linewidth]{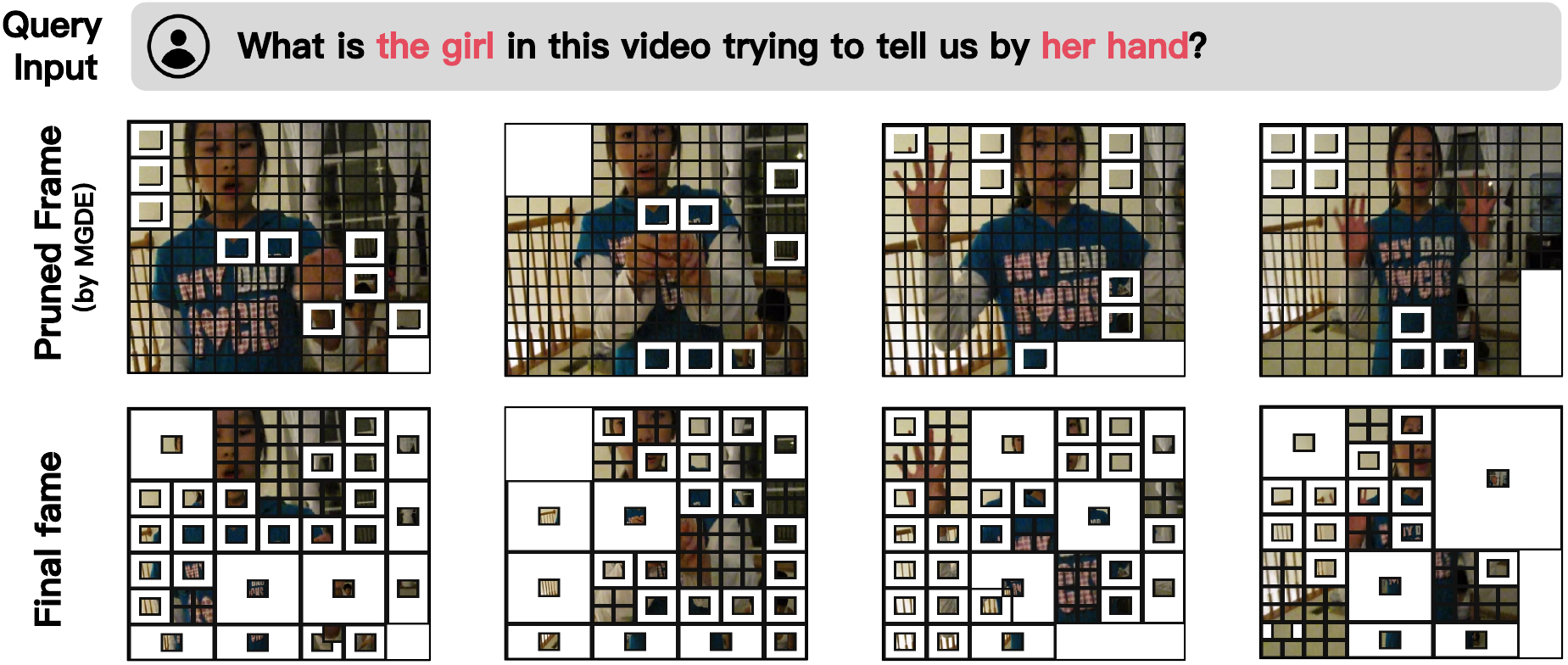}
\end{figure}

\subsection{Qualitative Analysis on InternVL3.5}

The two examples compare InternVL3.5-8B with and without FoveaTree, showing the sampled frames, compressed representations, and model responses. In the yoga example, FoveaTree describes two women practicing yoga on the beach, while the full model misidentifies one participant as a man. In the emotion question, the answer changes from ``Unable to determine'' to ``Full of pressure.'' The retained regions cover the people and their poses, illustrating how compressed visual inputs can still support detailed descriptions and question answering.

\begin{figure}[H]
\centering
\includegraphics[width=0.9\linewidth]{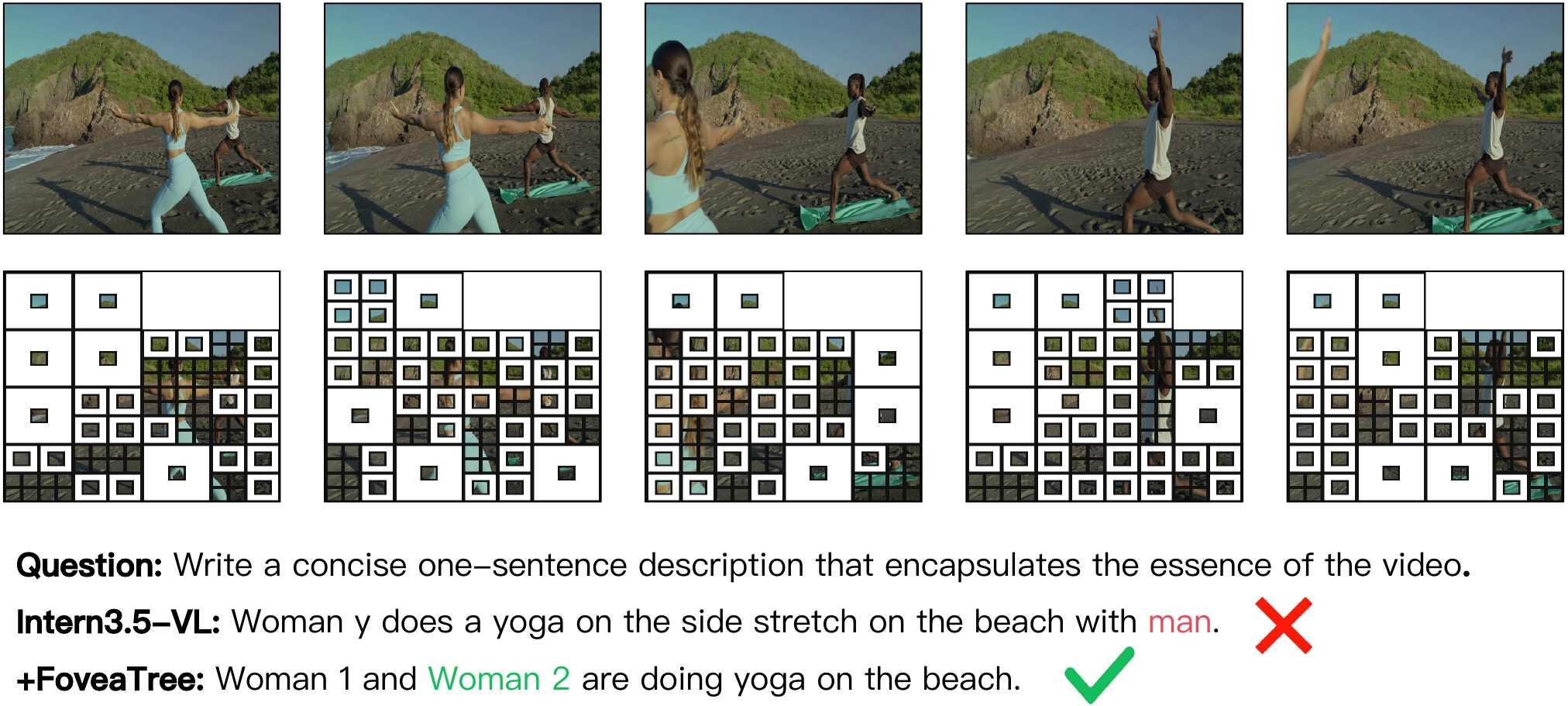}\par
\vspace{5pt}
\includegraphics[width=0.9\linewidth]{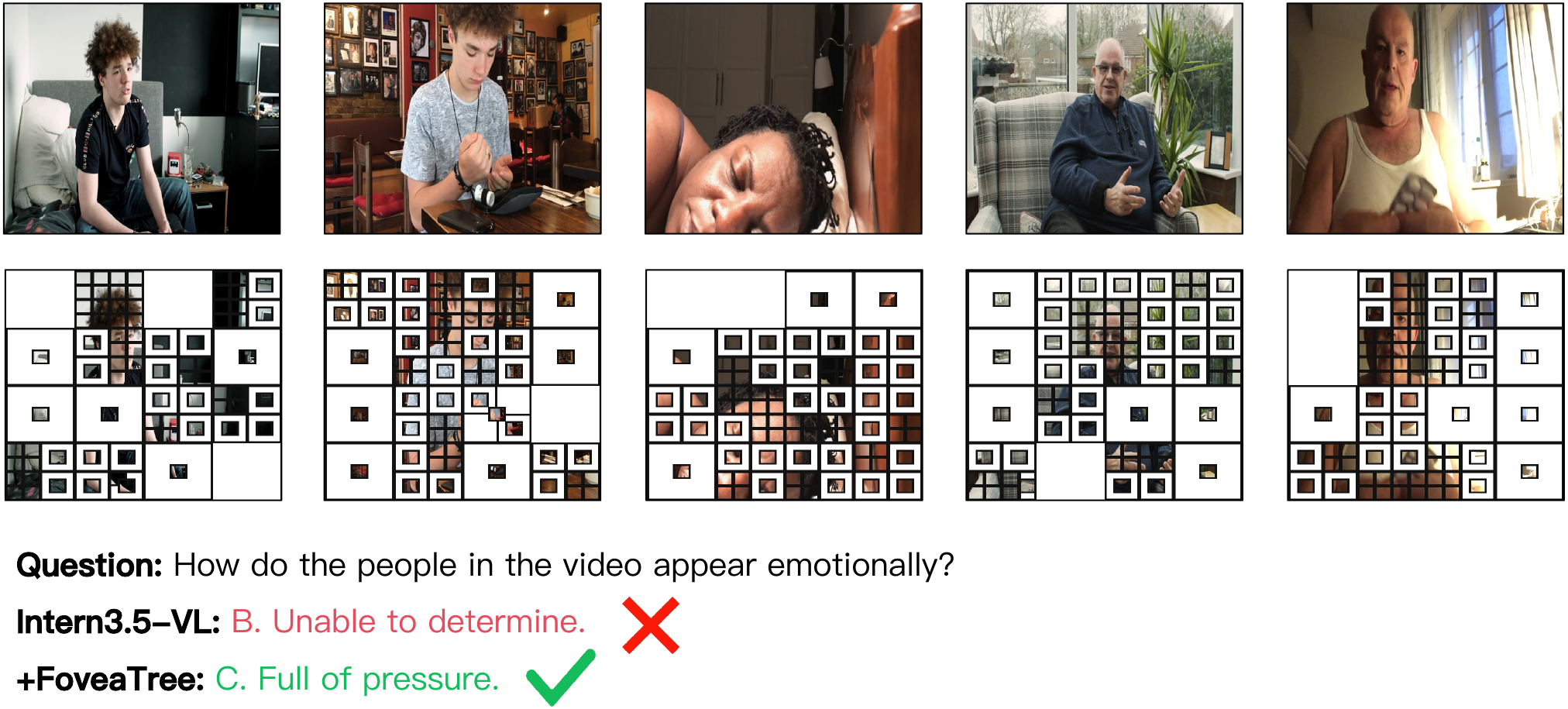}
\end{figure}

\subsection{Qualitative Analysis on Qwen3-VL}

The Qwen3-VL-8B examples cover numerical question answering and video captioning. For the financial-video question, FoveaTree selects the correct \$10 trillion difference between the two markets, while the full model selects \$5 trillion. In the balloon example, FoveaTree describes the girl tossing the balloon above her head and catching it, capturing the action sequence more precisely than the full model's description of throwing it away.

\begin{figure}[H]
\centering
\includegraphics[width=0.9\linewidth]{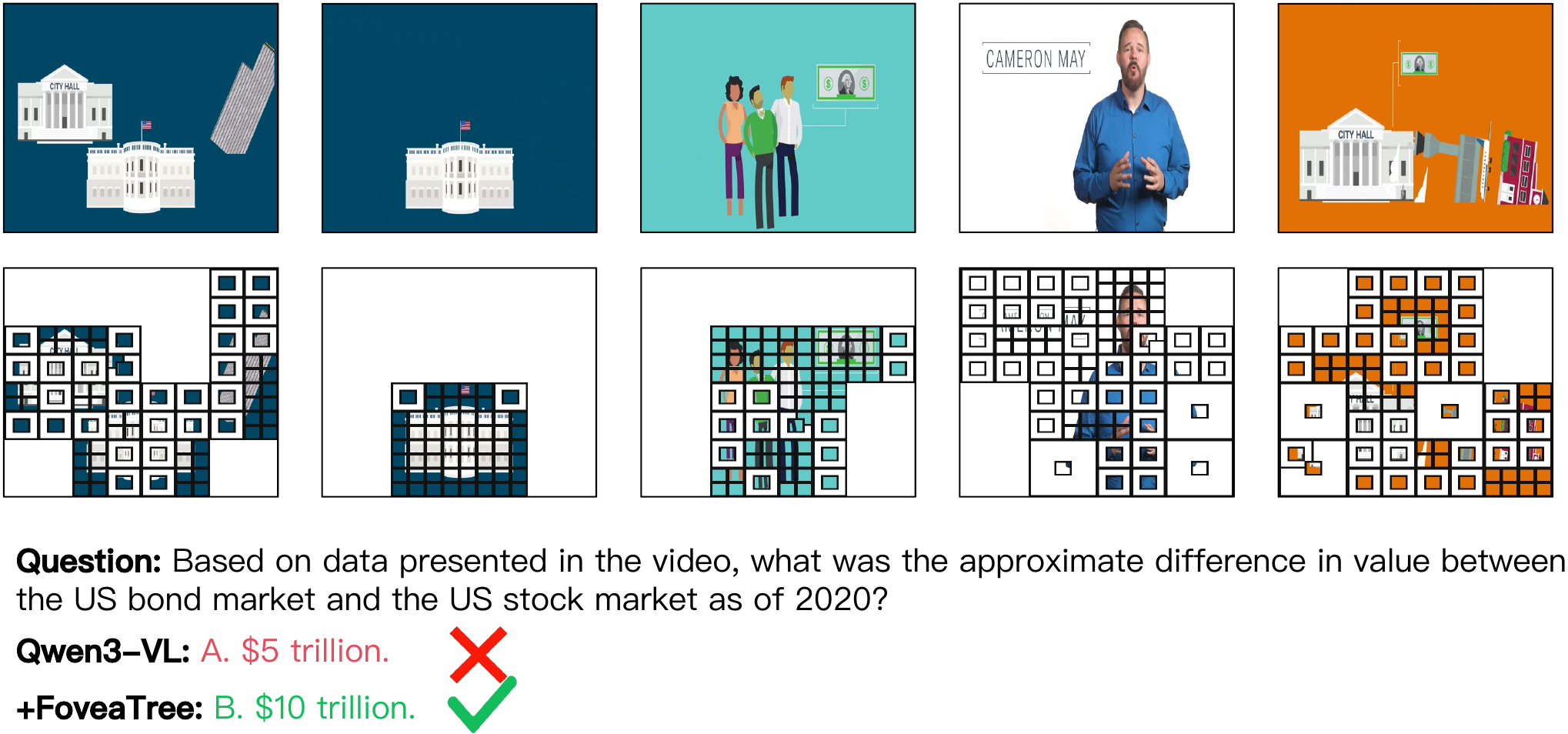}\par
\end{figure}

\begin{figure}[H]
\centering
\includegraphics[width=0.9\linewidth]{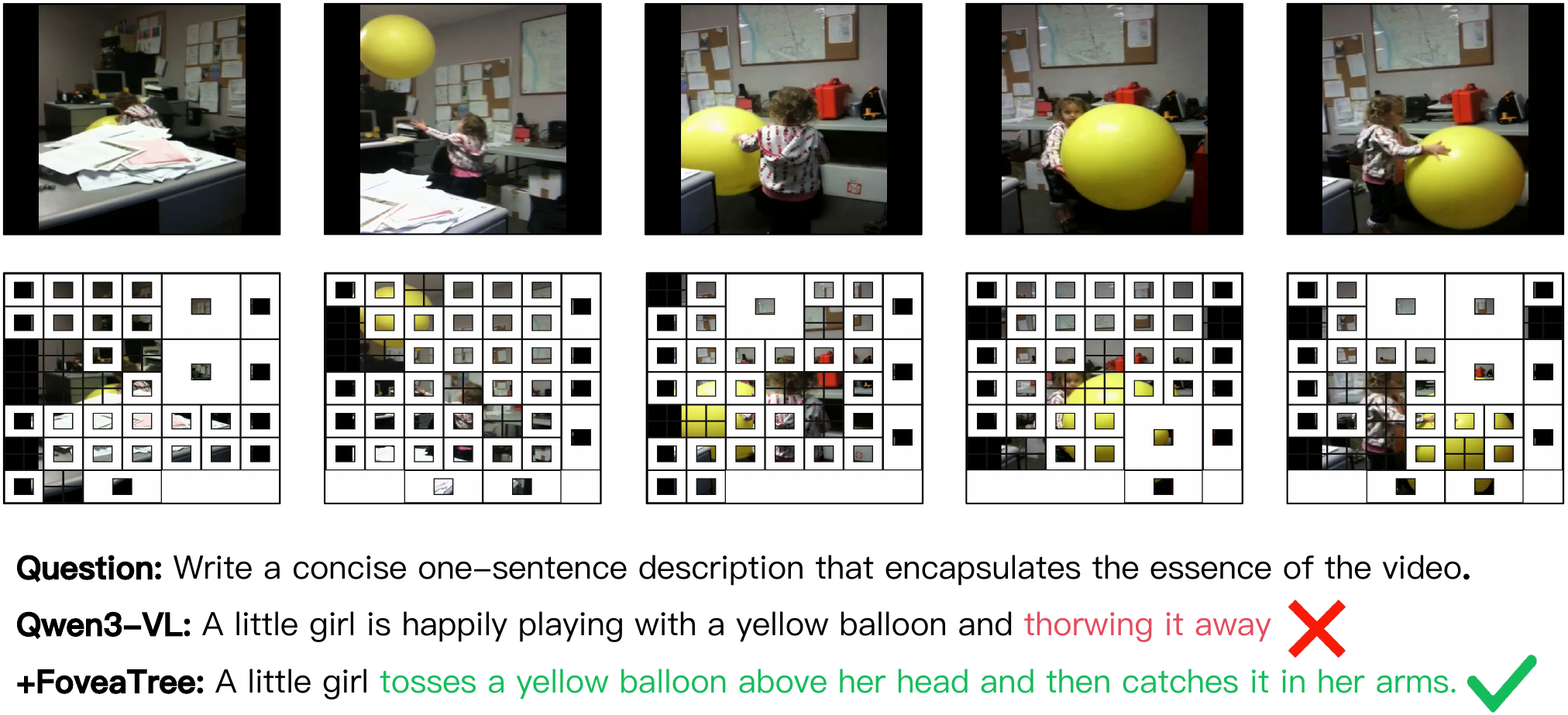}\par
\end{figure}

\end{document}